\documentclass[preprint,12pt]{elsarticle}
\usepackage{array}
\usepackage{amssymb}
\usepackage{amsmath}
\usepackage[utf8]{inputenc}
\usepackage{hyperref}
\usepackage{amsmath}
\usepackage{amsfonts}
\usepackage{amssymb}
\usepackage{geometry}
\usepackage{graphics}
\usepackage{graphicx}
\usepackage{bm}
\usepackage{tikz}
\usetikzlibrary{calc,hobby,arrows.meta,3d}
\usetikzlibrary{shadings}
\usepackage{booktabs}
\usepackage{multirow}
\usepackage{caption}
\usepackage{subcaption}
\usepackage{textcomp}
\usepackage{arydshln}
\usepackage{graphbox}
\hypersetup{colorlinks, citecolor=green, linkcolor=Red, urlcolor=Blue}
\biboptions{sort&compress}
\usepackage{float}
\usepackage{lineno}
\usepackage{rotating}
\usepackage{booktabs} 
\usepackage{dsfont}
\usepackage{rotating}
\usepackage{booktabs}
\usepackage{multirow}
\usepackage{makecell}
\usepackage{adjustbox}
\usepackage{rotating}
\usepackage{booktabs}
\usepackage{multirow}
\usepackage{adjustbox}
\usepackage{booktabs}

\begin{document}
\begin{frontmatter}

\title{Reconstructing and forecasting disease trajectories of patients with Alzheimer's disease using routine data in resource-constrained settings}

\author[mymainaddress1]{Ratnadeep Das}
\ead{ratnadeepdas97@gmail.com}

\author[mymainaddress2]{Atri Chatterjee}
\ead{chatterjee.atri@outlook.com}

\author[mymainaddress1,mymainaddress3]{Sitikantha Roy \corref{mycorrespondingauthor}}
\ead{sroy@am.iitd.ac.in}

\address[mymainaddress1]{Yardi School of Artificial Intelligence (ScAI), Indian Institute of Technology Delhi, Hauz Khas, Delhi, 110016, India}

\address[mymainaddress2]{Department of Neurology, Vardhman Mahavir Medical College and Safdarjung Hospital, Delhi, 110029, India}

\address[mymainaddress3]{Department of Applied Mechanics, Indian Institute of Technology Delhi, Hauz Khas, Delhi, 110016, India}
\cortext[mycorrespondingauthor]{Corresponding author}
\begin{abstract}
\textbf{Background and objective: }Alzheimer's disease is a progressive neurodegenerative disorder, and its progression varies substantially across patients. Patients often visit clinicians irregularly, making it difficult for clinicians to have a complete picture of a patient's disease trajectory. Existing work aims to forecast patients' future cognitive state, with minimal focus on reconstructing the state from past visits. Furthermore, the quantification of predictive uncertainty remains relatively underexplored, and most existing methods rely on costly modalities such as MRI, PET, and cerebrospinal fluid biomarkers, limiting their practical deployment in resource-limited settings. In this research, our primary objectives are: First, bidirectional prediction of cognitive scores from irregular visits to present the complete disease trajectory. Second, to enable interpolation and extrapolation capabilities to assist clinicians in informed prognostic decision making, and third, to provide a well-calibrated uncertainty estimate for all predictions, and finally, to achieve the objectives using the modalities available during routine visits. 

\noindent \textbf{Methods: }We propose a unified framework, GNOVA: A GRU-Neural ODE Variational Autoencoder. The architecture combines a Gated Recurrent Unit encoder and a Neural ODE decoder within a variational autoencoder framework. In our work, we forecast the CDR-SB and MMSE Scores. The GRU encoder allows for any number of inputs at any time point. The Neural-ODE decoder performs continuous estimation, allowing interpolation and extrapolation at any desired time point. The Variational autoencoder allows for uncertainty estimation in predictions. 

\noindent  \textbf{Results: }We worked with 1,727 patients from the Alzheimer's Disease Neuroimaging Initiative (ADNI) dataset over 10 years; the model achieved mean absolute errors of 1.35 and 2.28 for CDR-SB and MMSE scores, respectively, without requiring any neuroimaging or biomarker data. The model captured an average of 85\% of true CDR-SB values within predicted confidence intervals, with reduced coverage at longer extrapolation horizons. Feature-ablation studies revealed that age, BMI, and APOE4 status were strong predictors. 

\noindent  \textbf{Conclusion: }The proposed framework enables the reconstruction of incomplete patient histories and the anticipation of future cognitive states, thereby supporting individualized prognostic decision-making. Such work can encourage the development of models and frameworks that can be deployed in resource-constrained settings.

\end{abstract}



\begin{keyword}
Alzheimer's disease progression, disease trajectory modelling, neural ordinary differential equations, variational autoencoder, uncertainty quantification, resource-constrained settings
\end{keyword}

\end{frontmatter}


\section{Introduction}\label{sec:intro}
Alzheimer's disease (AD) is a progressive neurodegenerative disease that damages brain cells over time, causing an irreversible decline in cognitive abilities \cite{Evans2019}. According to the World Health Organization, in 2021, approximately 57 million people were suffering from dementia worldwide, with 10 million new cases each year. Globally, AD is the seventh leading cause of death and costs around \$1.3 trillion annually, accounting for around 70\% of all dementia cases \cite{who_dementia_2024}. Currently incurable, its progression can only be delayed through timely clinical intervention.
\par Initially, AD could be detected only at the autopsy. However, advances in technology have enabled clinicians to rely on modalities such as neuroimaging, fluid tests, and cognitive assessments to diagnose and monitor AD \cite{Blennow2017}. AD is characterized by abnormal $\beta-$amyloid ($A$$\beta$42) plaque accumulation and tau protein tangles within the brain  \cite{Zhang2024}. Positron Emission Tomography (PET) imaging and cerebrospinal fluid (CSF) amyloid-$\beta$42 ($A$$\beta$42) levels can detect these hallmarks with high sensitivity \cite{Schaap2024, Lowe2024}. However, these tests are costly, invasive, and require skilled personnel, making routine longitudinal follow-up impractical in most clinical settings. As a result, clinicians often rely primarily on cognitive assessment tools to track disease progression. In a recent study, Schaap et al. found a strong correlation between cognitive scores and amyloid positivity. Results showed that MMSE scores begin declining approximately 0.2 years after a patient becomes amyloid-positive, while CDR-SB scores worsen around 1.4 years after onset  \cite{Schaap2024}. This underscores the clinical value of tracking cognitive scores to track the disease progression. In this study, we aim to predict the trajectory of two widely used cognitive assessments: the Clinical Dementia Rating-Sum of Boxes (CDR-SB, range 0 to 18), which integrates cognitive and functional domains \cite{Tzeng2022}, and the Mini-Mental State Examination (MMSE, range 0 to 30), which evaluates orientation, attention, memory, language, and visuospatial skills \cite{Folstein1975}. 

\par AD progression varies substantially across patients. Among mild cognitive impairment (MCI) patients, Mouchet et al. found that approximately 21\% experience rapid progression, 22\% experience slow progression, and 57\% experience no progression over four years  \cite{Mouchet2021}. This study highlights the uncertainty in disease progression and the importance of continuously tracking scores to understand its nature. However, in practical scenarios, clinicians often have a weak historical context of the disease due to the patient’s irregular visits or visits only when the condition worsens. This makes it hard for clinicians to distinguish between patients with varied progression rates, thus delaying their treatment decisions. This is precisely the scenario that motivates the need for bidirectional trajectory modeling: a model that can not only forecast future cognitive states but also retrospectively reconstruct past ones from sparse, irregularly spaced observations.  
\par In the past decade, advancements in artificial intelligence have led researchers to develop computational tools for modeling AD progression. Initial efforts relied on traditional machine learning algorithms — support vector regression  \cite{Lei2020}, tree-based methods \cite{Jiang2021}, gradient boosting methods \cite{Devanarayan2024, Tabarestani2020}, and Gaussian processes (GPs) \cite{Peterson2017, Puri2022}. The introduction of deep learning enabled researchers to incorporate neuroimaging modalities and capture non-linear relationships for forecasting cognitive scores \cite{Liu2020, Morar2020, Yuan2024}. However, these methods typically forecast over short time horizons of three to five years and do not capture the inherent temporal dependencies in longitudinal data. Although GPs can model a 10-year trajectory, they scale poorly, with existing studies limited to around 100 patients  \cite{Peterson2017, Puri2022}. Recurrent neural networks (RNNs) and their gated variants — long short-term memory (LSTM) and gated recurrent unit (GRU) \cite{Goodfellow2016, Cho2014} naturally address temporal modeling of sequential data. Researchers have employed RNN variants to improve prediction performance and imputation capabilities: MinimalRNN \cite{Nguyen2020} and DeepRNN \cite{Jung2021} for individualized predictions, and Time\_LSTM \cite{Liang2021} for multi-task learning. Mukherji et al. and Morar et al. used many-to-one architectures for long-term prediction and effective temporal dependency capture \cite{Mukherji2022, Morar2023}, while encoder-decoder architectures  \cite{Cho2014} were later used to address variable sequence lengths in classification tasks \cite{Poonam2023, Poonam2024}. Das et al. found in their experiments that GRUs performed slightly better at predicting ADAS-13 and CDR-SB scores than RNNs and LSTMs \cite{Das}. Also, GRUs have comparable performance and are faster than LSTMs \cite{Goodfellow2016, chung2014empirical}. Statistical approaches such as linear mixed effects models naturally accommodate repeated measures and irregular visit schedules, but they are constrained to point estimates, assume linearity, and cannot construct bidirectional trajectories or interpolate predictions at arbitrary unobserved time points.  
\par Recently, a promising approach to handling irregularly sampled time-series data has emerged through neural ordinary differential equations (Neural ODEs). Unlike traditional discrete models that process data at fixed intervals, Neural ODEs model system dynamics continuously \cite{Chen2018}, allowing cognitive score progression to be represented as a smooth trajectory that can be evaluated at any desired time point. This is particularly valuable for AD, where progression is inherently continuous but clinical observations occur at variable and often widely spaced intervals. Jeong et al. demonstrated this potential using a GRU-ODE architecture for modeling the trajectory of cognitive scores \cite{Jeong2024}. While their approach captured continuous dynamics effectively, it remained deterministic and did not quantify prediction uncertainty. 

\par  GRU-based models capture temporal dependencies well, but operate only at discrete observed time points. Neural ODE-based models enable continuous dynamics but lack principled uncertainty estimation. Variational autoencoders alone provide probabilistic outputs but do not model continuous temporal evolution. To our best knowledge, no existing approach operates solely on routine clinical data and combines all three capabilities within a unified framework to predict disease trajectory for AD. In this work, we introduce GNOVA, which addresses these gaps by combining the strengths of GRU for sequential temporal encoding, Neural ODE for continuous latent trajectory modeling, a variational autoencoder for principled uncertainty estimation, and a dedicated encoder for static covariates. The resulting model enables clinicians to reconstruct incomplete patient histories, forecast future cognitive states at any desired time point, and obtain calibrated confidence intervals around all predictions — using only routine demographic and clinical data collected during standard visits, without requiring neuroimaging or biomarker testing. This makes the framework deployable in resource-constrained settings where advanced diagnostic infrastructure is unavailable. 

\section{Methods and materials}\label{sec:methods}
\subsection{Data}
\subsubsection{Dataset and Participants}
In our work, we used the Alzheimer’s Disease Neuroimaging Initiative (ADNI) dataset \cite{Mueller2005}. It is accessible at \url{https://adni.loni.usc.edu/}. ADNI includes diverse modalities, including MRI and PET scans, demographic and routine clinical information, and cognitive assessment scores. For our study, we included participants who had a baseline assessment (denoted by `bl'), a one-year follow-up (`m12'), and at least one additional follow-up visit within 10 years of baseline (`m120'). A total of 1727 patients met these criteria. Among them, 977 patients were male (56.57\%), and 750 were female (43.43\%). To build the model, we had selected the following features: demographic data (age, gender [Male/Female], years of education, BMI), hypertension status (Yes/No), diagnosis status (cognitive normal [CN], mild cognitive impairment [MCI], Dementia [AD]), APOE4 gene status (non-carrier, heterozygous, homozygous) and cognitive scores (CDR-SB and MMSE). We selected features that are routinely collected during standard clinical visits, requiring no neuroimaging or biomarker infrastructure. Table ~\ref{tab:feature_description} provides the complete description of all input features, their types, encodings, and resulting dimensionality.



\subsubsection{Data preprocessing}
The categorical features in our dataset were one-hot encoded, and the continuous variables were scaled by their maximum value. To handle missing values, we used forward filling to impute the values. There are several techniques for handling missing values. However, as noted in a study by Das et al. \cite{Das}, the forward filling technique performs closely with other methods in terms of predictive accuracy, and is simple to implement and understand. All features were imputed, except the target features (CDR-SB and MMSE scores) and Age, which could be imputed by adding or subtracting the visit difference. We did not impute the target variables (CDR-SB and MMSE), as imputing the prediction targets would yield an incorrect estimate of model performance during evaluation. Visit codes from months after baseline were converted to fractional years. For example, baseline ('bl') was encoded as 0.0, 'm06' as 0.5, 'm12' as 1.0, and so on, with the maximum visit code of 'm120' encoded as 10.0. The patient counts and descriptive statistics for annual and mid-year visits are presented in Tables~\ref{tab:yearly_summary} and~\ref{tab:mid_year_summary}, respectively.

\begin{table}[htbp]
  \centering
  \footnotesize
  \renewcommand{\arraystretch}{1.1}
  \setlength{\tabcolsep}{4pt}
  \begin{adjustbox}{width=\textwidth,center}
    \begin{tabular}{l*{11}{c}}
      \toprule
      \multirow{2}{*}{\textbf{Features}} & \multicolumn{11}{c}{\textbf{Time Points}} \\
      \cmidrule(l){2-12}
      & \textbf{0} & \textbf{1} & \textbf{2} & \textbf{3} & \textbf{4} & \textbf{5} & \textbf{6} & \textbf{7} & \textbf{8} & \textbf{9} & \textbf{10} \\
      \midrule
      \textbf{Count} & 1727 & 1727 & 1332 & 851 & 690 & 371 & 329 & 250 & 195 & 133 & 113 \\
      \midrule
      \multicolumn{12}{l}{\textit{Continuous Variables (Mean ± SD):}} \\
      \addlinespace[0.2em]
      BMI & 26.77±5.35 & 26.71±5.36 & 26.44±5.58 & 26.27±6.06 & 26.1±7.29 & 26.28±6.31 & 25.86±6.23 & 25.67±6.74 & 26.12±6.1 & 25.43±6.73 & 25.91±6.8 \\
      Education & 16.0±2.76 & 16.0±2.76 & 15.99±2.79 & 16.08±2.75 & 16.12±2.71 & 16.1±2.81 & 16.02±2.81 & 15.99±2.86 & 15.97±2.74 & 16.08±2.74 & 16.06±2.65 \\
      Age & 73.65±7.22 & 74.64±7.22 & 75.59±7.22 & 76.38±7.11 & 76.88±6.8 & 78.09±6.77 & 79.46±6.62 & 80.21±6.45 & 80.9±6.63 & 81.63±6.45 & 82.19±6.21 \\
      CDR-SB & 1.61±1.72 & 2.13±2.56 & 2.46±3.12 & 2.23±2.84 & 2.01±3.04 & 2.47±3.34 & 2.3±3.58 & 2.45±3.71 & 2.32±3.39 & 2.4±3.64 & 1.81±2.92 \\
      MMSE & 27.23±2.62 & 26.47±3.92 & 26.19±4.59 & 26.59±4.26 & 26.82±4.25 & 26.16±4.91 & 26.4±5.0 & 26.54±4.77 & 26.48±4.75 & 26.2±4.8 & 27.04±4.16 \\
      \midrule
      \multicolumn{12}{l}{\textit{Categorical Variables (Count):}} \\
      \addlinespace[0.2em]
      \multicolumn{12}{l}{\textbf{Diagnosis}} \\
      \quad CN & 486 & 496 & 422 & 233 & 269 & 119 & 144 & 97 & 79 & 46 & 48 \\
      \quad MCI & 926 & 809 & 531 & 414 & 276 & 161 & 113 & 100 & 70 & 53 & 46 \\
      \quad AD & 315 & 422 & 379 & 204 & 145 & 91 & 72 & 53 & 46 & 34 & 19 \\
      \addlinespace[0.3em]
      \multicolumn{12}{l}{\textbf{APOE4}} \\
      \quad Non & 923 & 923 & 725 & 488 & 394 & 221 & 209 & 156 & 127 & 90 & 74 \\
      \quad Hetero & 626 & 626 & 477 & 286 & 246 & 123 & 97 & 77 & 56 & 35 & 32 \\
      \quad Homo & 178 & 178 & 130 & 77 & 50 & 27 & 23 & 17 & 12 & 8 & 7 \\
      \addlinespace[0.3em]
      \multicolumn{12}{l}{\textbf{Hypertension}} \\
      \quad No & 883 & 883 & 700 & 448 & 361 & 203 & 180 & 140 & 113 & 78 & 69 \\
      \quad Yes & 844 & 844 & 632 & 403 & 329 & 168 & 149 & 110 & 82 & 55 & 44 \\
      \addlinespace[0.3em]
      \multicolumn{12}{l}{\textbf{Gender}} \\
      \quad Female & 750 & 750 & 586 & 362 & 303 & 162 & 141 & 112 & 81 & 62 & 51 \\
      \quad Male & 977 & 977 & 746 & 489 & 387 & 209 & 188 & 138 & 114 & 71 & 62 \\
      \bottomrule
    \end{tabular}
  \end{adjustbox}
  \caption{Data summary at yearly visits (0-10). CN = Cognitively Normal; MCI = Mild Cognitive Impairment; AD = Alzheimer's Disease; CDR-SB = Clinical Dementia Rating-Sum of Boxes; MMSE = Mini-Mental State Examination.}
  \label{tab:yearly_summary}
\end{table}

\begin{table}[htbp]
  \centering
  \footnotesize
  \renewcommand{\arraystretch}{1.1}
  \setlength{\tabcolsep}{6pt}
  \begin{adjustbox}{width=\textwidth,center}
    \begin{tabular}{l*{9}{c}}
      \toprule
      \multirow{2}{*}{\textbf{Features}} & \multicolumn{9}{c}{\textbf{Time Points}} \\
      \cmidrule(l){2-10}
      & \textbf{1.5} & \textbf{2.5} & \textbf{3.5} & \textbf{4.5} & \textbf{5.5} & \textbf{6.5} & \textbf{7.5} & \textbf{8.5} & \textbf{9.5} \\
      \midrule
      \textbf{Count} & 327 & 21 & 9 & 24 & 76 & 75 & 84 & 49 & 40 \\
      \midrule
      \multicolumn{10}{l}{\textit{Continuous Variables (Mean ± SD):}} \\
      \addlinespace[0.2em]
      BMI & 25.78±4.4 & 28.7±5.91 & 27.57±3.88 & 28.29±4.71 & 27.17±5.49 & 28.17±4.54 & 26.86±5.84 & 26.85±3.88 & 26.72±5.76 \\
      Education & 15.77±2.99 & 15.62±2.67 & 15.44±2.51 & 16.88±2.25 & 16.68±2.55 & 16.75±2.5 & 16.8±2.75 & 17.22±2.43 & 17.32±2.46 \\
      Age & 76.16±7.14 & 76.55±8.29 & 74.24±8.91 & 77.47±7.08 & 77.32±7.01 & 78.19±6.68 & 79.29±6.71 & 79.8±7.17 & 81.36±6.14 \\
      CDR-SB & 2.59±1.95 & 2.24±3.69 & 1.94±3.64 & 1.98±2.68 & 2.03±3.67 & 1.72±3.28 & 1.36±2.59 & 1.57±2.82 & 1.16±1.94 \\
      MMSE & 25.89±3.51 & 25.33±4.02 & 26.67±5.57 & 27.58±2.73 & 27.14±4.31 & 27.39±3.76 & 27.3±3.86 & 27.69±2.87 & 27.85±2.52 \\
      \midrule
      \multicolumn{10}{l}{\textit{Categorical Variables (Count):}} \\
      \addlinespace[0.2em]
      \multicolumn{10}{l}{\textbf{Diagnosis}} \\
      \quad CN & 10 & 8 & 5 & 8 & 44 & 34 & 37 & 16 & 22 \\
      \quad MCI & 224 & 7 & 2 & 13 & 20 & 29 & 37 & 26 & 14 \\
      \quad AD & 93 & 6 & 2 & 3 & 12 & 12 & 10 & 7 & 4 \\
      \addlinespace[0.3em]
      \multicolumn{10}{l}{\textbf{APOE4}} \\
      \quad Non & 151 & 12 & 5 & 15 & 48 & 48 & 52 & 33 & 28 \\
      \quad Hetero & 136 & 7 & 3 & 8 & 23 & 24 & 29 & 15 & 10 \\
      \quad Homo & 40 & 2 & 1 & 1 & 5 & 3 & 3 & 1 & 2 \\
      \addlinespace[0.3em]
      \multicolumn{10}{l}{\textbf{Hypertension}} \\
      \quad No & 164 & 10 & 3 & 11 & 39 & 46 & 48 & 33 & 19 \\
      \quad Yes & 163 & 11 & 6 & 13 & 37 & 29 & 36 & 16 & 21 \\
      \addlinespace[0.3em]
      \multicolumn{10}{l}{\textbf{Gender}} \\
      \quad Female & 117 & 9 & 5 & 6 & 34 & 30 & 36 & 18 & 16 \\
      \quad Male & 210 & 12 & 4 & 18 & 42 & 45 & 48 & 31 & 24 \\
      \bottomrule
    \end{tabular}
  \end{adjustbox}
  \caption{Data summary at mid-year visits (1.5-9.5). CN = Cognitively Normal; MCI = Mild Cognitive Impairment; AD = Alzheimer's Disease; CDR-SB = Clinical Dementia Rating-Sum of Boxes; MMSE = Mini-Mental State Examination.}
  \label{tab:mid_year_summary}
\end{table}

\subsubsection{Experimental Design}
After preprocessing, we divided the dataset into two groups \textemdash $D1$ and $D2$. $D1$ contained patients with annual visit data for visit codes 0-7. This dataset was split 70/10/20 for training, validation, and testing, respectively. We limited $D1$ to visit 7 as patient counts drop substantially beyond that point, which would introduce excessive noise into training. 
\par $D2$ contained participants' baseline, first follow-up data (visit codes 0 and 1) alongside remaining semi-annual visits: 1.5, 2.5, 3.5, 4.5, 5.5, and 6.5 for testing interpolation capabilities, and visits: 7.5, 8, 8.5, 9, 9.5, 10 for extrapolation testing. $D2$ was used exclusively for testing, with no involvement in training or validation.

\subsection{Model Architecture}
\subsubsection{Gated Recurrent Unit (GRU)}
Gated Recurrent Units (GRUs) are the gated variant of recurrent neural networks (RNNs) that can process longitudinal data and capture temporal dependencies \cite{Cho2014}. GRUs address the vanishing gradient problem of vanilla RNNs by introducing update and reset gates that control which information is retained and discarded across time steps. Given a sequence of inputs $\{x^{(i)}\}_{i=1}^{N}$, it produces hidden states $\{h^{(i)}\}_{i=1}^{N}$ which holds the temporal information of the sequence. 

\subsubsection{Variational Autoencoder (VAE)}
Autoencoders are a type of neural network architecture that learns a compressed representation of inputs, enabling it to reconstruct the input from that representation. Variational Autoencoders (VAEs) extend it by incorporating probabilistic modeling through variational Bayesian inference, enabling the learning of structured latent representations while maintaining generative capabilities \cite{Kingma2013}. Rather than encoding inputs to a fixed point in latent space, a VAE encodes them to a distribution, parameterized by mean $\mu$ and standard deviation $\sigma$, from which latent samples are drawn. This distributional encoding is what enables the model to produce confidence intervals around its predictions rather than deterministic point estimates, which is critical for clinical applications where overconfident predictions can mislead treatment decisions. A brief discussion on VAE architecture is given in appendix ~\ref{appendixB}

\subsubsection{Neural Ordinary Differential Equation}

Neural Ordinary Differential Equations (NODEs) are a class of deep learning models that replace the traditional discrete sequence of hidden layers with a continuous-depth framework governed by ordinary differential equations \cite{Chen2018}. In a discrete residual network, the evolution of the hidden state from $t$ to $t+1$ is given by
\begin{equation}
h_{t+1} = h_t + f(h_t, \theta)
\end{equation}
Where $\theta$ is the parameter of the function $f$.
In Neural ODE, instead of stacking a fixed number of layers, we treat the hidden state $h(t)$ evolving under an ordinary differential equation given as
\begin{equation}
\frac{dh(t)}{dt} = f(h(t), t; \theta)
\end{equation}

Where $f$ is a neural network parameterized by $\theta$. The Neural ODE solves the ODE over an interval $[t, t + 1]$ to give the next state as shown in equation \ref{eq:integral}.

\begin{equation}
h_{t+1} = h_t + \int_{t}^{t+1} f(h(\tau), \tau; \theta)d\tau
\label{eq:integral}
\end{equation}

In practice, we approximate this using an appropriate numerical ODE solver. The notation is given in equation \ref{eq:odesolver1}  

\begin{equation}
h_{t+1} = ODESolve(h_t,t,t+1,\theta)
\label{eq:odesolver1}
\end{equation}

\subsubsection{Final Architecture}

\begin{figure}[H]  
    \centering
    \includegraphics[width=0.7\textwidth]{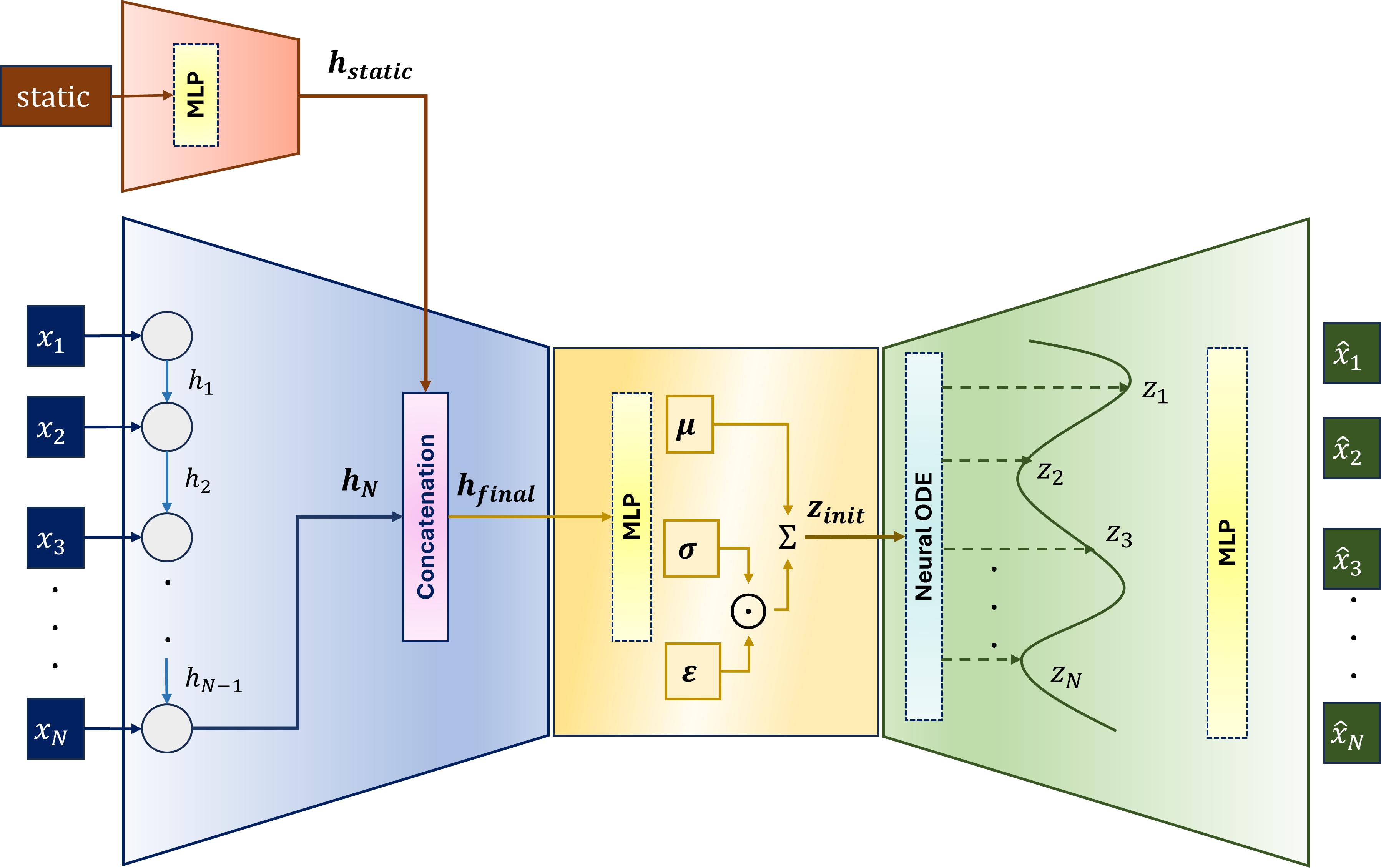}
    \caption{GNOVA Architecture}
    \label{fig:Figure_3}
\end{figure}

The complete GNOVA architecture is given in Figure \ref{fig:Figure_3}

The complete architecture consists of four blocks. The details of each block are given below. 
\begin{enumerate}[(i)]
    \item \textbf{Static Encoder Block: }It consists of a multi-layer perceptron (MLP) that nonlinearly transforms the 14-dimensional baseline static feature vector $x_{\text{static}}$ into a fixed-dimensional representation $h_{\text{static}}$ (\textit{encoder\_hidden\_dim}), using ReLU activation.
    
    \item \textbf{Sequence Encoder Block: }It consists of a GRU that processes the longitudinal sequence of input cognitive scores $\{x_i\}_{i=1}^{N}$ and produces hidden states $\{h_i\}_{i=1}^{N}$. The final hidden state $h_{\text{N}}$ is concatenated with $h_{\text{static}}$ to form the joint representation $h_{\text{final}}$ of dimension 2*\textit{encoder\_hidden\_dim}, integrating both the patient's temporal history of cognitive scores and their baseline clinical profile. We explored several schemes for the GRU hidden states, including averaging all hidden states $\{h_i\}_{i=1}^{N}$ and an attention-weighted sum of them. However, we found that the concatenation of the final state 
    performed better, as shown in the ablation study (section \ref{sec:ablation_study}). 
    
    \item \textbf{Latent Dynamics block: }It consists of an MLP (\textit{latent\_dim}) that maps $h_{\text{final}}$ to the mean and the variance of the latent distribution. The reparameterization trick is
    applied to form the initial latent variable $z_{\text{init}}$ given by equation \ref{eq:zinit}.
    
    \begin{equation}
    z_{\text{init}} = \mu + \sigma \odot \varepsilon, \quad \varepsilon \sim \mathcal{N}(0,I)
    \label{eq:zinit}
    \end{equation}
    
    \item \textbf{Decoder block: }In this block, the Neural ODE framework receives $z_{\text{init}}$ as input and integrates the latent trajectory continuously from t=0 to all desired prediction time points as given by equation \ref{eq:odesolver}.
    
    \begin{equation}
    z_{t+1} = ODESolve(z_{init},0,..,t+1,\theta)
    \label{eq:odesolver}
    \end{equation}

    The resulting latent variables $\{z_i\}_{i=1}^{N}$ are passed through an MLP (\textit{decoder\_hidden\_dim}) with Tanh activation to output the mean and variance of the predicted cognitive scores $\{\hat{x}_i\}_{i=1}^{N}$. In all experiments, we set $N=7$.
\end{enumerate}

\subsection{Training and Validation}
\subsubsection{Loss Function}
We employ the standard VAE loss function given in equation \ref{eq:elbo}. The function consists of reconstruction loss ($\mathcal{L}_{rec}$), which is the mean squared error (MSE) of the actual inputs and predicted values. It also has a regularization term ($\mathcal{L}_{reg}$) to prevent the model from predicting infinite uncertainty \cite{Kendall2017}. The total loss consisted of the reconstruction loss
($\mathcal{L}_{rec}$), KL divergence loss ($\mathcal{L}_{KL}$) ensuring the latent distribution remains close to the prior, and the regularizer loss ($\mathcal{L}_{reg}$), given by equation \ref{eq:totalloss} 
\begin{equation}
\mathcal{L}_{total} = \mathcal{L}_{rec} + \beta\mathcal{L}_{KL} + \lambda\mathcal{L}_{reg}
\label{eq:totalloss}
\end{equation}

Where $\beta$ and $\lambda$ are the hyperparameters used to control the weightage of the loss terms. Expanding each term, we have the function given by
\begin{equation}
\mathcal{L} = \frac{1}{2}\left(\frac{(x_i - \hat{x}_i)^2}{\sigma^2}\right) + \frac{\beta}{2}(1 + \log\sigma_j^2 - \sigma_j^2 - \mu_j^2) + \frac{\lambda}{2}(\log\sigma^2)
\end{equation}

Here, $\sigma^2$ is the predictive variance from the decoder, and $\sigma_j^2$ is the posterior variance for latent dimension $j$ from the encoder. The reconstruction loss uses mean squared error.

\subsubsection{Experimental setup and hyperparameters}

We implemented our model in PyTorch. For hyperparameter optimization, we used the Optuna framework \cite{akiba2019optuna} with 100 trials and selected parameters that minimized the validation loss. Table \ref{tab:hyperparameters} shows the search ranges and optimal values for each cognitive score model.
\begin{table}[h]
\centering
\footnotesize
\begin{tabular}{lccc}
\hline
\textbf{Parameter} & \textbf{Search Range} & \textbf{CDR-SB} & \textbf{MMSE} \\
\hline
encoder\_hidden\_dim & [16, 256] & 32 & 154 \\
latent\_dim & [16, 256] & 57 & 14 \\
decoder\_hidden\_dim & [16, 256] & 67 & 60 \\
learning\_rate & [0.0001, 0.1] & 0.0051 & 0.0067 \\
$\beta$ & [0, 1] & 0.691 & 0.202 \\
$\lambda$ & [0, 1] & 0.222 & 0.164 \\
\hline
\end{tabular}
\caption{Hyperparameter search space and optimal values for CDR-SB and MMSE models}
\label{tab:hyperparameters}
\end{table}

Models were trained for 500 epochs with a batch size of 16 using the Adam optimizer. We performed 5-fold cross-validation to ensure robust performance estimates. The ODE solver used default tolerance settings, which provided a good balance between accuracy and computational efficiency.

\subsubsection{Training and Inference Mechanism}

The ADNI dataset \cite{Mueller2005} is a prospective, well-organized cohort study in which subjects undergo regular study visits. Many subjects, however, might have dropped out due to a variety of reasons, and it gets reflected in the table ~\ref{tab:yearly_summary} and~\ref{tab:mid_year_summary} where the sample sizes become smaller with time. However, this missingness cannot be used for predictions in the experiment. Predicting these values cannot be used to assess the model's goodness, as we do not have ground-truth values. Hence, we use only the available time points to predict, mimicking a real-life scenario of irregular visits. All participants in the dataset who met the criteria discussed in the Data section above had baseline (visit 0) and one-year follow-up (visit 1) visits. We used only the input from those two time points for training. The other visits, except 0 and 1, are used for training as targets. During testing, however, we give any number of inputs at any desired time point. For example, we train the model with input time points 0,1, but we test the model with input time points 2,5, as shown in the Result sections later. This depicts a real-life scenario in which a clinician might need to predict a patient's disease trajectory at random time points. During training and inference, we follow the below masking strategy 

\begin{itemize}
\item \textbf{Training: }For each patient, we created a binary mask where 1 indicates the presence of data, and 0 indicates missing visits. The missing values were filled with zeros and were not considered during the calculation of the loss. This ensured that the model learned only from actual observations and maintained a consistent input dimension. The static encoder block was fed with the features at visit 0. 
\item \textbf{Inference: }During inference, we passed the values at the input time point (e.g, 0 and 1, or 2 and 5, etc.) and set the other values to 0. The masking was applied after we generated the outputs. The values we wanted to predict were masked to 1, and the input time points were set to 0. The static encoder block was fed the features of the earliest time point. (0 and 2 respectively as per the earlier example). 
\end{itemize}

\section{Results}\label{sec:data}
\subsection{Results on Dataset \texorpdfstring{$D1$}{D1} - Forward Prediction \& Retrospective Imputation }
We evaluated the model performance using 5-fold cross-validation on the test set (20\% of $D1$). To assess the model's flexibility with respect to input configuration, we tested the trained model across all possible pairs of input time points. Here, we present two scenarios: the forward prediction using the earliest available visits and retrospective imputation using the latest available visits. The results for all remaining input combinations are in \ref{appendixC}. The two scenarios are - 
\begin{itemize}
    \item Forward prediction: The model predicts cognitive scores at years 2 through 7 from baseline and first follow-up observations (Visit 0 and 1). This represents the clinical scenario in which the model forecasts from a minimal early history.
    \item Retrospective imputation: The model reconstructs earlier time points (visits 0 through 5) from late-trajectory observations (visits 6 \& 7). This is relevant in scenarios where a patient visits late in their disease course, and the clinician needs to reconstruct the earlier history.
\end{itemize}
The results of our experiment are given in the Table~\ref{tab:combined_prediction_results} for CDR-SB and MMSE.

\begin{table}[H]
\centering
\scriptsize
\begin{tabular}{cccccc}
\noalign{\hrule height 1pt}
\multicolumn{6}{c}{Input - 0,1} \\
\noalign{\hrule height 1pt}
\multirow{2}{*}{Target Time Point} & \multirow{2}{*}{No of patients} & \multicolumn{2}{c}{CDR-SB} & \multicolumn{2}{c}{MMSE} \\
\cline{3-4} \cline{5-6}
& & MAE & RMSE & MAE & RMSE \\
\noalign{\hrule}
2 & 266.4$\pm$2.6 & 0.8775$\pm$0.0582 & 1.4104$\pm$0.0977 & 1.7553$\pm$0.0926 & 2.4917$\pm$0.2479 \\
3 & 170.2$\pm$8.5 & 1.0519$\pm$0.1490 & 1.6605$\pm$0.2340 & 1.8831$\pm$0.0794 & 2.8474$\pm$0.1500 \\
4 & 138.0$\pm$10.1 & 1.1476$\pm$0.1590 & 1.9041$\pm$0.3843 & 1.9974$\pm$0.1711 & 3.0687$\pm$0.3372 \\
5 & 74.2$\pm$11.3 & 1.5342$\pm$0.1907 & 2.4050$\pm$0.3465 & 2.3953$\pm$0.2666 & 3.7750$\pm$0.5666 \\
6 & 65.8$\pm$3.9 & 1.5182$\pm$0.2726 & 2.4302$\pm$0.3884 & 2.3411$\pm$0.4662 & 3.6658$\pm$0.9266 \\
7 & 50.0$\pm$5.2 & 1.7380$\pm$0.1810 & 2.6138$\pm$0.2114 & 2.3993$\pm$0.2938 & 3.5895$\pm$0.7028 \\
\noalign{\hrule}
\multicolumn{6}{c}{Input - 6,7} \\
\noalign{\hrule}
\multirow{2}{*}{Target Time Point} & \multirow{2}{*}{No of patients} & \multicolumn{2}{c}{CDR-SB} & \multicolumn{2}{c}{MMSE} \\
\cline{3-4} \cline{5-6}
& & MAE & RMSE & MAE & RMSE \\
\noalign{\hrule}
0 & 345.4$\pm$0.5 & 0.6541$\pm$0.0282 & 0.9928$\pm$0.0651 & 1.4869$\pm$0.0191 & 1.8779$\pm$0.0294 \\
1 & 345.4$\pm$0.5 & 1.0598$\pm$0.0397 & 1.6429$\pm$0.0452 & 1.9658$\pm$0.0966 & 2.7555$\pm$0.1659 \\
2 & 266.4$\pm$2.6 & 1.3176$\pm$0.1299 & 2.0253$\pm$0.1943 & 2.2156$\pm$0.1869 & 3.2138$\pm$0.4271 \\
3 & 170.2$\pm$8.5 & 1.4281$\pm$0.1486 & 2.2117$\pm$0.3029 & 2.3075$\pm$0.0706 & 3.5007$\pm$0.2372 \\
4 & 138.0$\pm$10.1 & 1.3536$\pm$0.1317 & 2.2926$\pm$0.3255 & 2.2310$\pm$0.1523 & 3.4543$\pm$0.4783 \\
5 & 74.2$\pm$11.3 & 1.4183$\pm$0.1577 & 2.2640$\pm$0.4311 & 2.2737$\pm$0.2298 & 3.5549$\pm$0.5834 \\
\noalign{\hrule height 1pt}
\end{tabular}
\caption{Results for CDR-SB and MMSE predictions with different input configurations}
\label{tab:combined_prediction_results}
\end{table}

For forward prediction, the model attained a competitive MAE of 0.8775$\pm$0.0582 at year 2, progressively rising to 1.7380$\pm$0.1810 at year 7 for CDR-SB. MMSE predictions exhibited a similar pattern, with MAE of 1.7553$\pm$0.0926 at year 2 increasing to 2.3993$\pm$0.2938 at year 7. For retrospective imputation, the model achieved its best performance in baseline reconstruction, with MAEs of 0.6541$\pm$0.0282 and 1.4869$\pm$0.0191 for CDR-SB and MMSE, respectively. 
\par Interestingly, retrospective predictions were not consistently better than forward predictions at time points close to the input. For example, at t=4, which is closer to the input window of t=6,7 than to t=0,1, retrospective MAE was 1.3536$\pm$0.1317 compared to forward MAE of 1.1476$\pm$0.1590 for CDR-SB. A similar pattern was observed in MMSE predictions.

\subsection{Results for Dataset \texorpdfstring{$D2$}{D2} - Interpolation \& Extrapolation}

To evaluate the model's capacity to generalize beyond its training conditions, we assessed the trained $D1$ model on the $D2$ dataset, which included semi-annual visits excluded from the training phase. The results are shown in Table~\ref{tab:combined_interpolation_extrapolation}. The model was trained on annual visits from t=0 to t=7 (utilizing $D1$, with t=0 and 1 as inputs and t=2 to 7 as targets) and evaluated on $D2$, where the interpolation range comprised t=1.5, 2.5, 3.5, 4.5, 5.5, and 6.5, while the extrapolation range included t=7.5, 8, 8.5, 9, 9.5, and 10. 

\begin{table}[H]
\centering
\scriptsize
\begin{tabular}{cccccc}
\noalign{\hrule height 1pt}
\multicolumn{6}{c}{Input - 0,1} \\
\noalign{\hrule height 1pt}
\multirow{2}{*}{Target Time Point} & \multirow{2}{*}{No of patients} & \multicolumn{2}{c}{CDR-SB} & \multicolumn{2}{c}{MMSE} \\
\cline{3-4} \cline{5-6}
& & MAE & RMSE & MAE & RMSE \\
\noalign{\hrule}
1.5 & 327 & 0.8213$\pm$0.0196 & 1.2072$\pm$0.0360 & 2.4645$\pm$0.1226 & 2.9777$\pm$0.1224 \\
2.5 & 21 & 0.8839$\pm$0.0535 & 1.4152$\pm$0.1341 & 1.5477$\pm$0.1475 & 1.8872$\pm$0.1926 \\
3.5 & 9 & 1.2427$\pm$0.1088 & 1.8039$\pm$0.2302 & 2.4706$\pm$0.4147 & 3.9910$\pm$0.3789 \\
4.5 & 24 & 0.7924$\pm$0.0534 & 1.2024$\pm$0.0852 & 1.5550$\pm$0.2849 & 1.9713$\pm$0.3049 \\
5.5 & 76 & 1.1848$\pm$0.0305 & 1.9289$\pm$0.0928 & 2.0996$\pm$0.1020 & 3.1698$\pm$0.1100 \\
6.5 & 75 & 1.5545$\pm$0.1257 & 2.3691$\pm$0.2341 & 2.1250$\pm$0.1650 & 3.0381$\pm$0.1569 \\
\noalign{\hrule}
7.5 & 84 & 1.6426$\pm$0.1584 & 2.6580$\pm$0.3153 & 2.3049$\pm$0.0892 & 3.3058$\pm$0.1660 \\
8 & 195 & 1.7151$\pm$0.0797 & 2.5508$\pm$0.0857 & 2.4463$\pm$0.1927 & 3.9443$\pm$0.4208 \\
8.5 & 49 & 2.0075$\pm$0.3286 & 2.9461$\pm$0.5548 & 2.4025$\pm$0.1656 & 3.2603$\pm$0.2442 \\
9 & 133 & 2.2122$\pm$0.1059 & 3.2339$\pm$0.1007 & 3.1661$\pm$0.2932 & 4.9525$\pm$0.6427 \\
9.5 & 40 & 1.8383$\pm$0.2443 & 2.4084$\pm$0.4861 & 2.5838$\pm$0.8208 & 3.4534$\pm$0.7474 \\
10 & 113 & 2.1530$\pm$0.2051 & 2.9800$\pm$0.2474 & 2.8090$\pm$0.5531 & 4.6702$\pm$0.7766 \\
\noalign{\hrule height 1pt}
\end{tabular}
\caption{Interpolation and extrapolation results for CDR-SB and MMSE}
\label{tab:combined_interpolation_extrapolation}
\end{table}

The interpolation performance was predominantly robust, especially at closer time periods. The exception was the MMSE's performance at t=1.5, which was not as good as CDR-SB's. Interestingly, we noticed a correlation between sample size and forecast stability. At t=3.5, with just 9 patients available, the CDR-SB MAE variance was $\pm$0.1088, and for MMSE, it was $\pm$0.4147, both exhibiting much greater variability than at prior interpolation time points. This instability, most likely, arises from the limited sample size. 
\par In the case of extrapolation, too, our model performed well at near-time points, and its predictive performance declined steadily as the distance from the training window increased. However, it is expected behavior for any model operating beyond its observed range. CDR-SB MAE rose from 1.6426$\pm$0.1584 at t=7.5 to 2.1530$\pm$0.2051 at t=10. However, it is essential to recognize that the variance in CDR-SB scores is considerable at these time points. The standard deviation of actual CDR-SB scores at t=9 and t=10, as indicated in Table 1, is approximately 3.6 and 2.9, respectively. Consequently, the model's mean absolute error of approximately 2.1–2.2 indicates a prediction error within the data's inherent variability. Moreover, the number of patients at these remote time points is much lower (n=113 at t=10), which makes these estimates less reliable statistically. MMSE extrapolation indicated a significant deterioration, with MAE attaining 3.1661$\pm$0.2932 at t=9, implying that MMSE trajectories are more challenging to extrapolate over extended periods compared to CDR-SB trajectories.

\subsection{Observations within confidence intervals}
To evaluate the reliability of the model's uncertainty estimates, we computed the percentage of true values falling within the predicted 95\% confidence intervals at each time point. Results are shown in Table~\ref{tab:combined_observations_in_CI}.

\begin{table}[H]
\scriptsize
\centering
\begin{tabular}{cccc}
\noalign{\hrule height 1pt}
\multicolumn{4}{c}{Input - 0,1} \\
\noalign{\hrule height 1pt}
\multirow{2}{*}{Target Time Point} & \multirow{2}{*}{No of patients} & \multicolumn{2}{c}{No of observations in CI (in \%)} \\
\cline{3-4}
& & CDR-SB & MMSE \\
\noalign{\hrule}
1.5 & 327 & 94.50$\pm$0.86 & 85.26$\pm$3.47 \\
2 & 266.4$\pm$2.6 & 92.41$\pm$0.59 & 88.82$\pm$1.74 \\
2.5 & 21 & 90.48$\pm$3.01 & 91.43$\pm$4.67 \\
3 & 170.2$\pm$8.5 & 90.61$\pm$2.97 & 89.87$\pm$2.52 \\
3.5 & 9 & 88.89$\pm$0.00 & 86.67$\pm$4.44 \\
4 & 138.0$\pm$10.1 & 89.35$\pm$1.45 & 90.39$\pm$1.32 \\
4.5 & 24 & 95.83$\pm$2.64 & 95.83$\pm$2.64 \\
5 & 74.2$\pm$11.3 & 86.13$\pm$4.16 & 85.29$\pm$3.47 \\
5.5 & 76 & 87.37$\pm$1.78 & 85.00$\pm$2.14 \\
6 & 65.8$\pm$3.9 & 84.69$\pm$2.15 & 82.14$\pm$8.59 \\
6.5 & 75 & 86.13$\pm$1.36 & 85.07$\pm$2.72 \\
7 & 50.0$\pm$5.2 & 81.48$\pm$4.12 & 77.20$\pm$6.63 \\
\noalign{\hrule}
7.5 & 84 & 87.38$\pm$3.33 & 79.52$\pm$2.95 \\
8 & 195 & 85.54$\pm$2.46 & 78.26$\pm$5.49 \\
8.5 & 49 & 84.49$\pm$3.79 & 75.92$\pm$4.36 \\
9 & 133 & 80.60$\pm$3.60 & 61.50$\pm$8.64 \\
9.5 & 40 & 91.00$\pm$4.64 & 66.50$\pm$16.78 \\
10 & 113 & 82.30$\pm$3.67 & 56.28$\pm$17.19 \\
\noalign{\hrule height 1pt}
\end{tabular}
\caption{Percentage of observations within the confidence interval (CI) for CDR-SB and MMSE predictions}
\label{tab:combined_observations_in_CI}
\end{table}

The coverage of CDR-SB was robust and consistent. Within the interpolation range (t=1.5 to t=6.5), coverage varied from 84.69\% to 95.83\%, with an average of about 90\% across folds. During the extrapolation interval (t=7.5 to t=10), coverage varied from 80.60\% to 91.00\%, with an average of roughly 85\%. The confidence intervals for CDR-SB are consistently well-calibrated in both interpolation and extrapolation, achieving over 80\% coverage at all assessed time periods. However, for MMSE predictions, the results were different. Within the interpolation range, MMSE coverage was comparable to CDR-SB, varying from 82.14\% to 95.83\% and averaging about 88\%. However, coverage dropped significantly beyond t=9, decreasing to 61.50\% at t=9 and 56.28\% at t=10. The model's uncertainty estimates are well-calibrated for MMSE within and near the training window; they become increasingly underestimated at long extrapolation horizons. 

\subsection{Effect of Input Configuration and Sequence Length}
We also conducted a comprehensive analysis to understand how the input time point influences prediction performance. First, we tested all single and paired time points as inputs to identify which observations were most informative for trajectory reconstruction. Second, we examined how prediction accuracy changed as we incrementally added more historical data points. Tables~\ref{tab:CDRSB_input_length} and~\ref{tab:CDRSB_input_target} show the results of the experiment for CDR-SB predictions and ~\ref{tab:MMSE_input_length} and~\ref{tab:MMSE_input_target} for MMSE predictions. 

As shown in Table~\ref{tab:CDRSB_input_length} and ~\ref{tab:MMSE_input_length}, prediction accuracy improved consistently as the number of input time points increased. Adding a third observation beyond visits 0 and 1 improved MAE across all subsequent time points by approximately 30\% on average for CDR-SB, demonstrating the model's effective use of additional historical context. The improvement was most visible at later time points, where the benefit of more historical information is greatest.
\par Table~\ref{tab:CDRSB_input_target} and~\ref{tab:MMSE_input_target} reveals a consistent pattern in single-input-pair performance: the best MAE for any target time point was generally achieved when the input time point immediately preceded it. For example, the best prediction at t=4 used the input at t=3 (MAE: 0.92), while at t=3, the input at t=2 (MAE: 0.88) for CDR-SB. Tables ~\ref{tab:CDRSB_two_inputs} and ~\ref{tab:MMSE_two_inputs} show the results for two input time points.

\subsection{Ablation Study} \label{sec:ablation_study}
We conducted a systematic ablation study to analyze two aspects: encoder aggregation strategy and static feature contribution. Results are reported in tables~\ref{tab:CDRSB_ablation} and~\ref{tab:MMSE_ablation} for CDR-SB and MMSE predictions, respectively.
\par For the static features aggregation, we tried three approaches: using the final hidden state (current model), the average of all hidden states, and an attention-weighted sum. We observed that no single approach performed best at all time points. However, on average, the final hidden state achieved better MAE for both CDR-SB (1.4464) and MMSE (2.2653), marginally outperforming the average hidden state approach (CDR-SB: 1.4491, MMSE: 2.3957) and the attention-based approach (CDR-SB: 1.4591, MMSE: 2.4309). 

\par The contribution of each static covariate was assessed by systematically removing it and observing the change in MAE. For CDR-SB predictions, age (MAE increase to 1.4891) and APOE4 status (1.4772) had the strongest influence, while MMSE as a covariate (1.4673) and BMI (1.4605) had moderate influence. Interestingly, removing hypertension (1.4148) and gender (1.4148) slightly improved CDR-SB predictions, suggesting these features may introduce noise rather than signal for CDR-SB trajectory modeling. However, this is not conclusive and requires further analysis to establish. For MMSE predictions, age (2.3982), APOE4 status (2.3880), diagnosis stage (2.3411), CDR-SB (2.3212), and BMI (2.3186) all had a strong influence. Surprisingly, removing years of education slightly improved MMSE predictions (2.2460 vs 2.2653). 

\subsection{Case studies}
\begin{figure}[H]  
    \centering
    \includegraphics[width=1\textwidth]{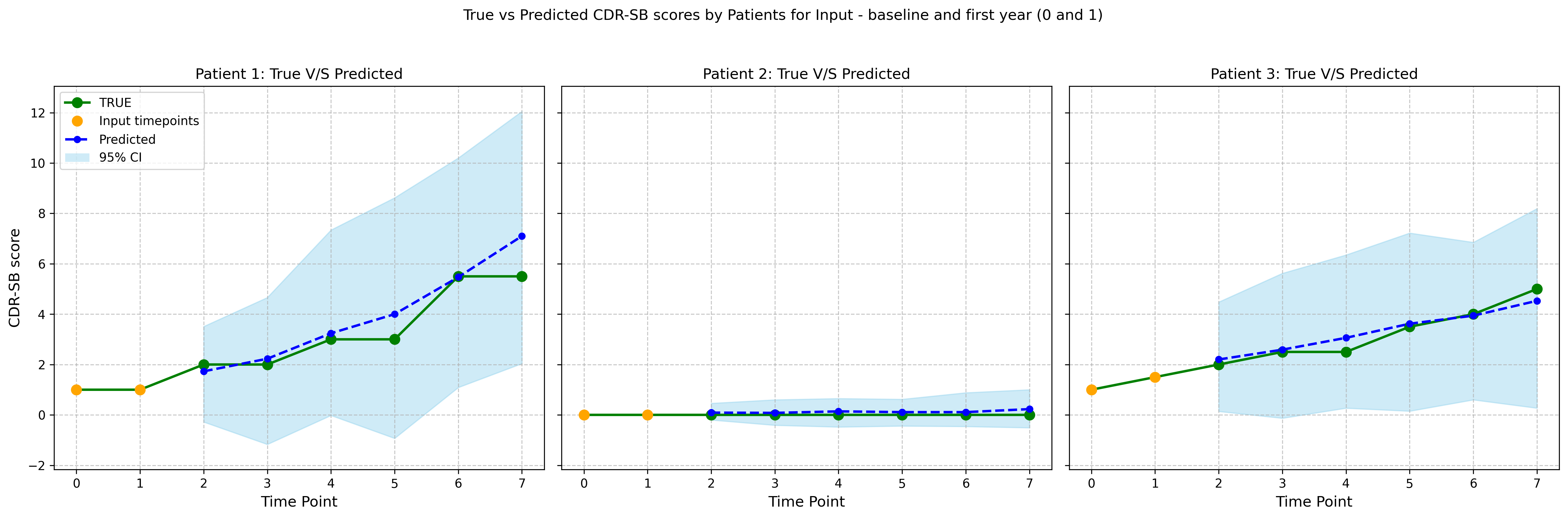}
    \caption{Trajectory of the patients with Input time points - 0,1}
    \label{fig:Figure_4}
\end{figure}

\begin{figure}[H]  
    \centering
    \includegraphics[width=1\textwidth]{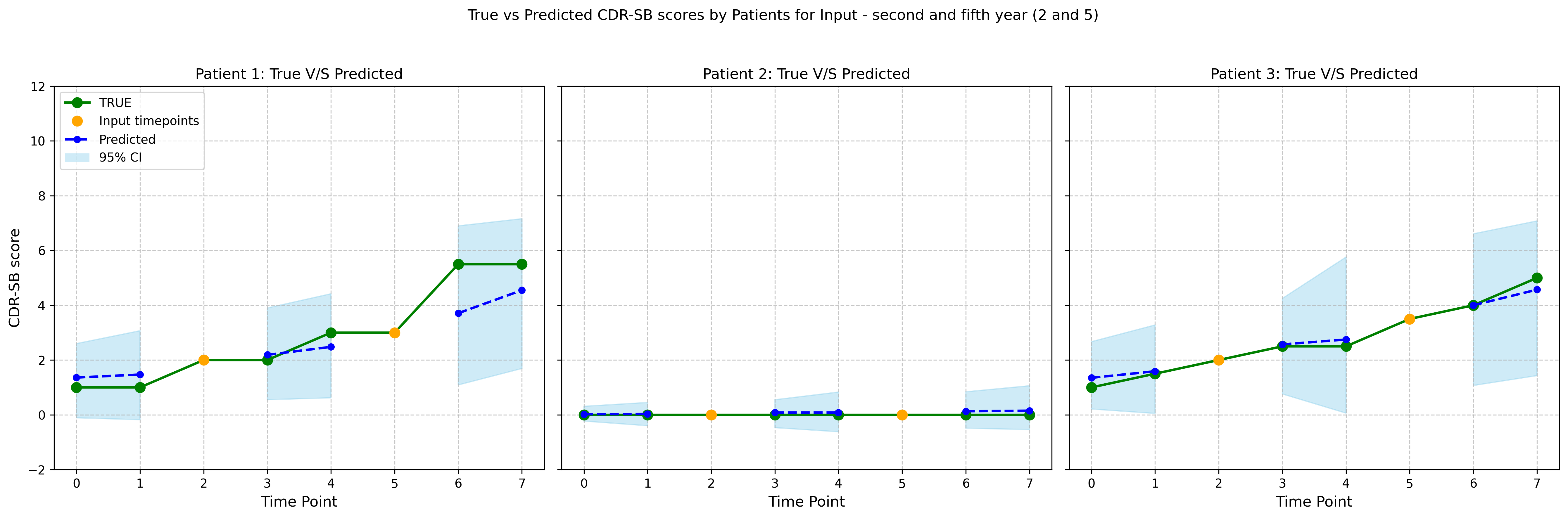}
    \caption{Trajectory of the patients with Input time points - 2,5}
    \label{fig:Figure_5}
\end{figure}

\begin{figure}[H]  
    \centering
    \includegraphics[width=1\textwidth]{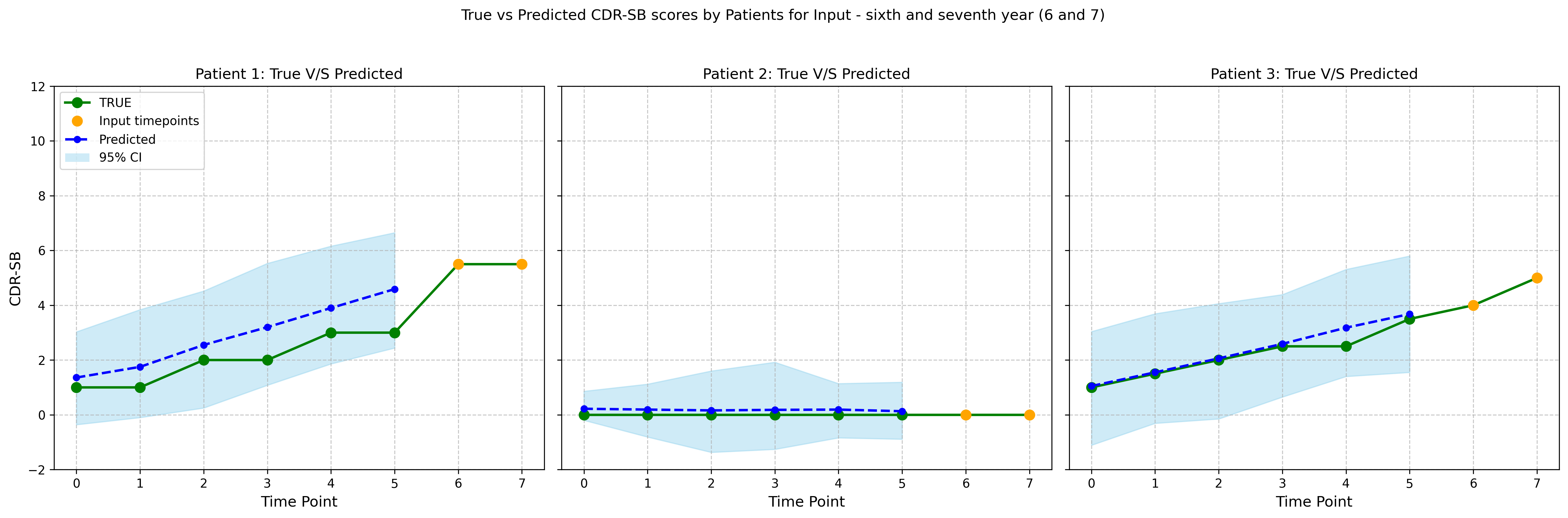}
    \caption{Trajectory of the patients with Input time points - 6,7}
    \label{fig:Figure_6}
\end{figure}

We present three case studies that represent clinically distinct progression characteristics, and we evaluate our model's predictions for those cases using three different input configurations: early visits (0,1), mid-trajectory visits (2,5), and late visits (6,7). Only patients with complete data across all seven annual time points were considered. The three patients chosen exhibit three distinct progression patterns frequently observed in Alzheimer's Disease: a patient with a sudden, nonlinear increase in CDR-SB ($P1$), a patient with stable, near-zero progression ($P2$), and a patient with consistently monotonically rising scores ($P3$). Figures ~\ref{fig:Figure_4}, ~\ref{fig:Figure_5}, and~\ref{fig:Figure_6} illustrate the confidence interval, predicted score, and true score for each patient and input configuration.

\par The model captured the progression trend well for patients $P2$ (stable) and $P3$ (monotonically increasing) at all the input configurations. Confidence intervals were appropriately narrow around the input time points and widened with increasing prediction distance, reflecting well-calibrated uncertainty.

\par For $P1$ (abrupt jumps), the model captured the sudden score jump from t=5 to t=6 successfully when early visits (0,1) were used as inputs. This is possibly because the model learned this sharp transition pattern during forward training. However, when mid-trajectory visits (2,5) were used, the model failed to anticipate the jump at t=6, producing a smoother trajectory that underestimates the severity of progression. For this input, it could capture the previous time point's trajectory, but failed to capture the future trajectory. In terms of point-value estimates, it was underestimated at all the time points. However, for the late inputs (6,7), the model could not reconstruct the preceding fall from t=6 to t=5 in terms of point values, but it well captured the progression trend. Nevertheless, this inability to capture the abrupt changes stems from the limitation of Neural ODE-based models: the continuous latent trajectory enforced by the ODE solver favors smooth dynamics and is inherently less suited to abrupt nonlinear transitions \cite{chen2020learning}. At t=5 and t=7 specifically, $P1$ shows notable discrepancy between true and predicted values across input configurations, and we acknowledge these as genuine prediction failures rather than artifacts of the evaluation. Clinically, this suggests the model may underperform among rapid progressors whose trajectories involve sudden deterioration. Extending the architecture to accommodate sharper trajectory dynamics is an important direction for future work.
\subsection{Comparison with previous works}
We compared our models' results with those of previous work at different time points using relevant metrics. Tables ~\ref{tab:cdrsb_comparison} and ~\ref{tab:mmse_comparison} show the comparison for CDR-SB and MMSE scores.

Our model demonstrates competitive performance across both CDR-SB and MMSE predictions and has the key advantage of using readily available clinical data rather than expensive neuroimaging and fluid modalities.

\par For CDR-SB predictions, our GNOVA model achieves strong results across multiple time points. Using only demographic and clinical data, our model outperforms previous methods at M12 (RMSE: 1.329, MAE: 0.849) and M24 (RMSE: 1.720, MAE: 1.133) that incorporated complex neuroimaging data. Even at M18 and M36, where Lei et al. \cite{Lei2022} report slightly better performance, our results remain competitive. Notably, their approach required MRI data and was tested on substantially smaller cohorts (M18: N=282; M36: N=50) than our larger testing set (M18: N=327; M36: N=170).

\par For MMSE predictions, our model demonstrates reasonable performance across multiple time points. In terms of MAE, our model achieves strong results at several key time points. At M12, we obtained an MAE of 1.864, closely competitive with Yuan et al. \cite{Yuan2024} (MAE: 1.69). Our performance remains consistent at M24 (MAE: 1.755) and M36 (MAE: 1.883), comparing favorably with Lei et al. \cite{Lei2022}. Our model's predictions at M18 had a slightly higher error. However, in our framework, prediction at M18 (t=1.5) is an interpolation task, while other methods specifically trained their models for M18 prediction. 

\par The RMSE results further validate our model's effectiveness. We achieve competitive performance through M36, with minimal deviation from best-reported results: M12 (Our: 2.56, Morar et al. \cite{Morar2023}: 2.17), M18 (Our: 2.98, Morar et al. \cite{Morar2023}: 2.18), M24 (Our: 2.49, Tabarestani et al. \cite{Tabarestani2020}: 2.38), and M36 (Our: 2.84, Tabarestani et al. \cite{Tabarestani2020}: 2.28). Our model achieves the best performance at M30. Post M36, RMSE increases moderately. Methods achieving lower errors post M36, such as Jung et al. \cite{Jung2021}, utilized less than 50\%  cohorts at each time point.  Notably, all the works used expensive imaging modalities (PET, MRI, CSF) limiting the practical applicability in a low-resource setting.

\begin{table}[htbp]
\centering
\scriptsize
\begin{tabular}{@{}p{3.5cm}p{2.2cm}p{2.5cm}p{2cm}p{2cm}@{}}
\toprule
\multicolumn{5}{c}{\textbf{CDR-SB (RMSE)}} \\
\midrule
\textbf{Proposed Works} & \textbf{Method} & \textbf{Modalities} & \textbf{M12} & \textbf{M24} \\
\midrule
Liu et al. \cite{Liu2020} & wiseDNN & MRI, CS & 2.008 & 2.334 \\
Our Model - Input (0,1) & GNOVA & Demo, CS & -- & \textbf{1.349} \\
Our model - Input 0 & GNOVA & Demo, CS & \textbf{1.329} & 1.720 \\
\bottomrule
\end{tabular}

\vspace{1.5em}

\begin{tabular}{@{}p{3.5cm}p{2.2cm}p{3cm}p{1.2cm}p{1.2cm}p{1.2cm}p{1.2cm}@{}}
\toprule
\multicolumn{7}{c}{\textbf{CDRSB (MAE)}} \\
\midrule
\textbf{Proposed Works} & \textbf{Method} & \textbf{Modalities} & \textbf{M12} & \textbf{M18} & \textbf{M24} & \textbf{M36} \\
\midrule
Lei et al. \cite{Lei2020} & SVR & MRI, CS & 0.965 & 0.861 & 1.396 & 0.816 \\
Lei et al. \cite{Lei2022} & IndRNN & MRI, CS & -- & \textbf{0.69} & 1.01 & \textbf{0.72} \\
Devanarayan et al. \cite{Devanarayan2024} & GB & MRI, Demo, CS & \shortstack{0.94\\0.99} & 1.15 & 1.35 & -- \\
Our Model - Input (0,1) & GNOVA & Demo, CS & -- & 0.821 & \textbf{0.878} & 1.052 \\
Our model - Input 0 & GNOVA & Demo, CS & \textbf{0.849} & -- & 1.133 & 1.336 \\
\bottomrule
\end{tabular}
\caption{Comparison of CDR-SB predictions with previous works at time points. CDR-SB: Clinical Dementia Rating Sum of Boxes; RMSE: Root Mean Square Error; MAE: Mean Absolute Error; MRI: Magnetic Resonance Imaging; CS: Cognitive Scores; Demo: Demographics; SVR: Support Vector Regression; IndRNN: Independently Recurrent Neural Network; GB: Gradient Boosting; GNOVA: Gated Recurrent Unit - NOVA architecture; M12/M18/M24/M36: 12/18/24/36 months from baseline.}
\label{tab:cdrsb_comparison}
\end{table}

\begin{sidewaystable}[htbp]
\centering

\scriptsize
\begin{tabular}{@{}lllcccc@{}}
\toprule
\multicolumn{7}{c}{\textbf{MMSE (MAE)}} \\
\midrule
\textbf{Proposed Works} & \textbf{Method} & \textbf{Modalities} & \textbf{M12} & \textbf{M18} & \textbf{M24} & \textbf{M36} \\
\midrule
Lei et al.\cite{Lei2020} & SVR & MRI, CS & 1.801 & 1.777 & 1.84 & 1.756 \\
Lei et al.\cite{Lei2022} & IndRNN & MRI, CS & -- & \textbf{1.74} & 1.92 & \textbf{1.56} \\
Yuan et al.\cite{Yuan2024} & MFSE-DRN & sMRI, CS, Demo, Others & \textbf{1.69} & -- & 1.86 & 2.03 \\
Our Model - Input 0,1 & GNOVA & Demo,CS & -- & 2.465 & \textbf{1.755} & 1.883 \\
Our model - Input 0 & GNOVA & Demo,CS & 1.864 & -- & 2.124 & 2.191 \\
\bottomrule
\end{tabular}

\vspace{2em}

\scriptsize
\begin{tabular}{@{}lllccccccccccccc@{}}
\toprule
\multicolumn{16}{c}{\textbf{MMSE (RMSE)}} \\
\midrule
\textbf{Paper} & \textbf{Method} & \textbf{Modalities} & \textbf{M12} & \textbf{M18} & \textbf{M24} & \textbf{M30} & \textbf{M36} & \textbf{M42} & \textbf{M48} & \textbf{M54} & \textbf{M60} & \textbf{M72} & \textbf{M84} & \textbf{M96} & \textbf{M108} \\
\midrule
Liu et al.\cite{Liu2020} & wiseDNN & MRI, CS & 3.128 & -- & 3.408 & -- & -- & -- & -- & -- & -- & -- & -- & -- & -- \\
Jung et al.\cite{Jung2021} & Deep-RNN & Volume, CS, Demo & 2.409 & -- & 2.483 & -- & 2.411 & -- & 2.64 & -- & \textbf{2.793} & \textbf{2.344} & \textbf{2.303} & \textbf{2.546} & \textbf{2.044} \\
Morar et al.\cite{Morar2020} & FCNN & Volume, CSF, PET, CS, Demo & 2.52 & 2.76 & 2.71 & 3 & 2.87 & 3.11 & 3.05 & 3.64 & 3.23 & -- & -- & -- & -- \\
Yuan et al.\cite{Yuan2024} & MFSE-DRN & sMRI, CS, Demo, Others & 2.48 & -- & 2.67 & -- & 3.02 & -- & -- & -- & -- & -- & -- & -- & -- \\
Tabarestani et al.\cite{Tabarestani2020} & GB-MTL & Volume, PET, CS, Demo, CSF & 2.24 & -- & \textbf{2.38} & -- & \textbf{2.28} & -- & \textbf{2.19} & -- & -- & -- & -- & -- & -- \\
Morar et al.\cite{Morar2023} & ST-LSTM & Volume, CSF, PET, CS, Demo & \textbf{2.17} & \textbf{2.18} & 2.61 & 2.52 & 2.71 & \textbf{2.67} & 3.17 & 3.01 & 2.9 & -- & -- & -- & -- \\
Our Model - Input 0,1 & GNOVA & Demo,CS & -- & 2.98 & 2.49 & \textbf{1.88} & 2.84 & 3.99 & 3.07 & \textbf{1.97} & 3.77 & 3.67 & 3.59 & 3.94 & 4.95 \\
Our model - Input 0 & GNOVA & Demo,CS & 2.56 & -- & 3.1 & -- & 3.28 & -- & 3.47 & -- & 4.05 & 4.01 & -- & -- & -- \\
\bottomrule
\end{tabular}

\caption{Comparison of MMSE predictions with previous works at time points. MMSE: Mini-Mental State Examination; MAE: Mean Absolute Error; RMSE: Root Mean Square Error; MRI: Magnetic Resonance Imaging; CS: Cognitive Scores; Demo: Demographics; sMRI: structural MRI; CSF: Cerebrospinal Fluid (A$\beta$,p-tau,t-tau); PET: Positron Emission Tomography (FDG, PIB, AV45); Volume - ventricles, hippocampus, fusiform gyrus, middle temporal gyrus, entorhinal cortex, and whole-brain; SVR: Support Vector Regression; IndRNN: Independently Recurrent Neural Network; MFSE-DRN: Multi-Feature Squeeze-and-Excitation Deep Residual Network; FCNN: Fully Connected Neural Network; GB-MTL: Gradient Boosting Multi-Task Learning; ST-LSTM: Single Task Long Short-Term Memory; M$X$: $X$ months from baseline}
\label{tab:mmse_comparison}

\end{sidewaystable}

\section{Discussion}\label{sec:results}
Our model achieved mean absolute errors of 1.35 and 2.28 for CDR-SB and MMSE, respectively, over a ten-year trajectory, without requiring any neuroimaging or biomarker modalities. A change of 1.0–2.0 points in CDR-SB is usually considered clinically significant in patients with mild-to-moderate Alzheimer’s disease. Such a change indicates a noticeable shift in cognitive or functional abilities and often leads clinicians to reassess the patient. In this context, the model’s average error (MAE) of 1.35 for CDR-SB lies within a clinically meaningful range. This is especially relevant for early predictions, where the errors are smaller and treatment decisions are more critical. In comparison, the MMSE score ranges from 0 to 30 but has known ceiling and floor effects due to different factors such as education \cite{FRANCOMARINA201072}. Patients at early or late stages of the disease tend to have scores clustered near the maximum or minimum values, which reduces the range of variation. As a result, an MAE of 2.28 reflects not only model error but also the inherent difficulty of predicting changes within this limited scale. At later time points, the predictions are more accurate when interpreted as indicative of trajectory direction rather than exact score values. The associated confidence intervals provide more useful information about the expected range of outcomes. 

\par The bidirectional prediction capability addresses a practical clinical need that is rarely addressed in existing literature. When a patient arrives with only a few inconsistent historical observations, a clinician must simultaneously reason about the patient's past and future. The majority of current methodologies treat forecasting and imputation as distinct tasks, necessitating the development a unified framework for each direction. Our architecture encompasses both within a singular, integrated framework. The model's retrospective capability to derive clinically plausible estimates is demonstrated by its ability to reconstruct baseline scores from late encounters, with an MAE of 0.6541 for CDR-SB baseline reconstruction. An intriguing finding was that forward predictions were occasionally more precise than retrospective ones at proximal time points. This may be attributed to the fact that forward neurodegeneration patterns are more consistently represented in the training data than reverse ones, given the natural directionality of disease progression. 

\par The model's probabilistic outputs provide a significant benefit compared to deterministic methods. In healthcare, a point prediction without an associated uncertainty estimate imposes the whole interpretative responsibility on the physician, to rely or not on the model's predictions. Well-calibrated confidence intervals effectively identify predictions characterized by significant model uncertainty. In our experiments, CDR-SB uncertainty estimates were consistently well-calibrated during the entire assessment period, achieving over 80\% coverage of actual targets within the projected 95\% confidence intervals at all time periods, including the extrapolation range. However, as we see in Table ~\ref{tab:combined_observations_in_CI}, the MMSE coverage at extrapolation did not perform well. The MMSE uncertainty estimates significantly deteriorated after t=9, with coverage decreasing to 61.50\% at t=9 and 56.28\% at t=10. This is a meaningful limitation, and in clinical use, the model's predictions at t=9 and t=10 for MMSE should be interpreted with awareness that the reported confidence intervals likely underestimate true uncertainty. Improving long-range uncertainty calibration for MMSE, potentially through recalibration techniques or ensemble approaches, is an important direction for future work.

\par The feature ablation study revealed few findings that are consistent with the established literature as well as some unexpected results. The strong influence of age and APOE4 status on both CDR-SB and MMSE predictions aligns with their well-documented roles as the strongest risk and progression factors in AD \cite{sando2008apoe,roses2006discovery,kim2009role}. The moderate influence of BMI is consistent with emerging evidence linking metabolic health to cognitive trajectory in AD \cite{cho2022association}. However, two unexpected findings necessitate further research. First, removing years of education marginally improved MMSE predictions, which is counterintuitive as many research has establish a relationship between years of education and MMSE \cite{xie2016distinct}. Second, removing hypertension and gender information slightly improved CDR-SB predictions. This is possibly because these features contribute more strongly to disease risk than to progression rate, and adds little predictive information beyond what is already provided by stronger covariates such as age and APOE4.

\par In the case studies, we noticed that the model performed well in capturing the trend of the progression. However, for retrospective imputation, it still estimate the point values correctly. This is expected, as the model is trained on forward prediction. Future work would focus on methods to improve the point-estimates. Also, the model shows a tendency to slightly overestimate cognitive scores. From a clinical perspective, this is preferable to underestimation. Overestimation is more likely to lead to additional checks and earlier intervention, whereas underestimation could delay necessary action. 

\section{Limitations}
We acknowledge several limitations of our model. Firstly, all experiments were conducted on the ADNI dataset, which, despite its size and longitudinal depth, is a North American prospective cohort with specific inclusion criteria that may not represent the full diversity of AD patients encountered in routine clinical practice. External validation on independent datasets, particularly those from clinical settings with genuinely irregular visit patterns, is necessary before any deployment and remains the highest priority for future work. 
\\ Secondly, the static treatment of many covariates, such as diagnosis stage, hypertension status, age, etc., is a deliberate simplification, and it is likely to limit prediction accuracy at longer horizons, where the baseline representation becomes increasingly outdated. In future work, we plan to introduce a dynamical block to encode such features. 
\\ Thirdly, the continuous latent trajectory enforced by the Neural ODE solver inherently favors smooth dynamics, making the model less suited to patients who exhibit abrupt transitions in cognitive scores, as illustrated by the P1 case study results. In future work, we can explore a second-order differential equation-based model of progression, taking into account where we can model the velocity and acceleration (trajectory and drifts) of the disease.  
\\ Fourth, a direct comparison of different architectural baselines, such as GRU-ODE without VAE and ODE-RNN was not included in this study, primarily because these architectures are very slow. While the internal ablation study and published literature comparison provide partial evidence for the proposed design choices, a controlled component-wise ablation would more rigorously establish the individual contribution of each architectural element and is planned for future work. 
\\ Finally, the model currently excludes neuroimaging and biomarker data by design. However, in practical situations, a certain modality may be present. This will limit the current architecture's ability to incorporate additional information. In the future, we plan to extend the framework to incorporate available biomarkers, while gracefully working with routine clinical data alone when they are absent, thereby broadening the model's utility across the full spectrum of resource availability in clinical settings.

\section{Conclusions}\label{sec:conclusion}
The proposed GNOVA framework demonstrates that disease trajectory modeling in Alzheimer's disease can be achieved using minimal routine clinical data as well, without any neuroimaging or biomarker infrastructure. The model integrates bidirectional prediction, continuous interpolation and extrapolation, and calibrated uncertainty quantification into a single framework, providing capabilities that are not available simultaneously in any current methodology. Its intentional reliance on low-cost, readily available data makes it especially appropriate for resource-limited healthcare environments that need decision-support tools. External validation on independent clinical datasets, comprehensive architectural ablation, and multimodal extensions remain important directions of future work. We hope this work encourages further research into practical, accessible tools for personalized dementia care.

\section*{CRediT authorship contribution statement}
\textbf{Ratnadeep Das}: Writing – original draft, Writing – review \& editing, Methodology, Data curation, Software, Visualization, Conceptualization, Formal analysis.
\textbf{Atri Chatterjee}: Writing – review \& editing, Supervision, Conceptualization, Formal analysis.
\textbf{Sitikantha Roy}: Writing – review \& editing, Supervision, Conceptualization, Formal analysis, Project administration.

\section*{Acknowledgments}
We acknowledge the subjects who participated in the Alzheimer’s Disease Neuroimaging Initiative (ADNI) and the team who made this work possible. The data collection and sharing were funded by the ADNI  (National Institutes of Health Grant U01 AG024904) and DOD ADNI (Department of Defense award number W81XWH-12-2-0012). ADNI is funded by the National Institute on Aging, the National Institute of Biomedical Imaging and Bioengineering, and through generous contributions from the following: AbbVie, Alzheimer’s Association; Alzheimer’s Drug Discovery Foundation; Araclon Biotech; BioClinica, Inc.; Biogen; Bristol-Myers Squibb Company; CereSpir, Inc.; Cogstate; Eisai Inc.; Elan Pharmaceuticals, Inc.; Eli Lilly and Company; EuroImmun; F. Hoffmann-La Roche Ltd and its affiliated company Genentech, Inc.; Fujirebio; GE Healthcare; IXICO Ltd.; Janssen Alzheimer Immunotherapy Research \& Development, LLC.; Johnson \& Johnson Pharmaceutical Research \& Development LLC.; Lumosity; Lundbeck; Merck \& Co., Inc.; Meso Scale Diagnostics, LLC.; NeuroRx Research; Neurotrack Technologies; Novartis Pharmaceuticals Corporation; Pfizer Inc.; Piramal Imaging; Servier; Takeda Pharmaceutical Company; and Transition Therapeutics. The Canadian Institutes of Health Research is providing funds to support ADNI clinical sites in Canada. Private sector contributions are facilitated by the Foundation for the National Institutes of Health (\url{www.fnih.org}). The grantee organization is the Northern California Institute for Research and Education, and the study is coordinated by the Alzheimer’s Therapeutic Research Institute at the University of Southern California. ADNI data are disseminated by the Laboratory for Neuro Imaging at the University of Southern California.

\section*{Data availability statement}
Data used in the preparation of this article were obtained from the Alzheimer’s Disease Neuroimaging Initiative (ADNI) database accessible in \url{adni.loni.usc.edu}. The ADNI, launched as a public-private partnership in 2003, was led by Principal Investigator Michael W. Weiner, MD. The goal of the dataset is to test whether modalities such as magnetic resonance imaging (MRI), positron emission tomography (PET), biological markers, and different clinical and neuropsychological scores can help us in identifying and predicting the progression of Alzheimer’s Disease.  As such, the investigators within the ADNI contributed to the design and implementation of ADNI and/or provided data but did not participate in the analysis or writing of this report. A complete listing of ADNI investigators can be found at: \url{http://adni.loni.usc.edu/wp-content/uploads/how_to_apply/ADNI_Acknowledgement_List.pdf}.

\section*{Ethics statement}
All human subjects involved in this dataset (clinical trial) provided written informed consent prior to their participation.

\section*{Declaration of Generative AI and AI-assisted technologies in the writing process}
During the preparation of this work, the authors used AI-based tools (including Google Gemini, Claude, ChatGPT, Grammarly, and QuillBot) to assist with grammar correction, linguistic refinement, and enhancement. After using these tools, the authors reviewed and edited the content as needed and take full responsibility for the content of the published article.

\section*{Funding sources}
This research did not receive any specific grant from funding agencies in the public, commercial, or not-for-profit sectors.

\appendix
\setcounter{table}{0}      
\setcounter{figure}{0}     
\section{Feature Description} \label{appendixA}
The table below gives the description of all the features

\begin{table}[htbp]
  \centering
  \scriptsize
  \renewcommand{\arraystretch}{1.1}
  \setlength{\tabcolsep}{4pt}
  \begin{adjustbox}{width=\textwidth,center}
    \begin{tabular}{lccc}
      \toprule
      \textbf{Feature} & \textbf{Type} & \textbf{Encoding} & \textbf{Dimensions} \\
      \midrule
      Age & Continuous & Scaled by maximum value & 1 \\
      Gender & Categorical & One-hot (Male (1) / Female (2)) & 2 \\
      Years of Education & Continuous & Scaled by maximum value & 1 \\
      BMI & Continuous & Scaled by maximum value & 1 \\
      Hypertension Status & Categorical & One-hot (Yes (1) / No (0)) & 2 \\
      Diagnosis & Categorical & One-hot (1 / 2 / 3) & 3 \\
      APOE4 Status & Categorical & One-hot (Non / Heterozygous / Homozygous) & 3 \\
      MMSE$^{\dagger}$ / CDR-SB$^{\ddagger}$ & Continuous & Scaled by maximum possible value & 1 \\
      \midrule
      \textbf{Total} & & & \textbf{14} \\
      \bottomrule
    \end{tabular}
  \end{adjustbox}
  \caption{Feature, type of data, encoding strategy, and resulting dimensionality for input vector. 1 = Cognitively Normal; 2 = Mild Cognitive Impairment; 3 = Dementia; BMI = Body Mass Index;
  $^{\dagger}$MMSE at baseline is included as a static covariate only in the CDR-SB prediction model.
  $^{\ddagger}$CDR-SB at baseline is included as a static covariate only in the MMSE prediction model.}
  \label{tab:feature_description}
\end{table}

\section{Variational Autoencoder}  \label{appendixB}
\begin{figure}[H]  
    \centering
    \includegraphics[width=0.5\textwidth]{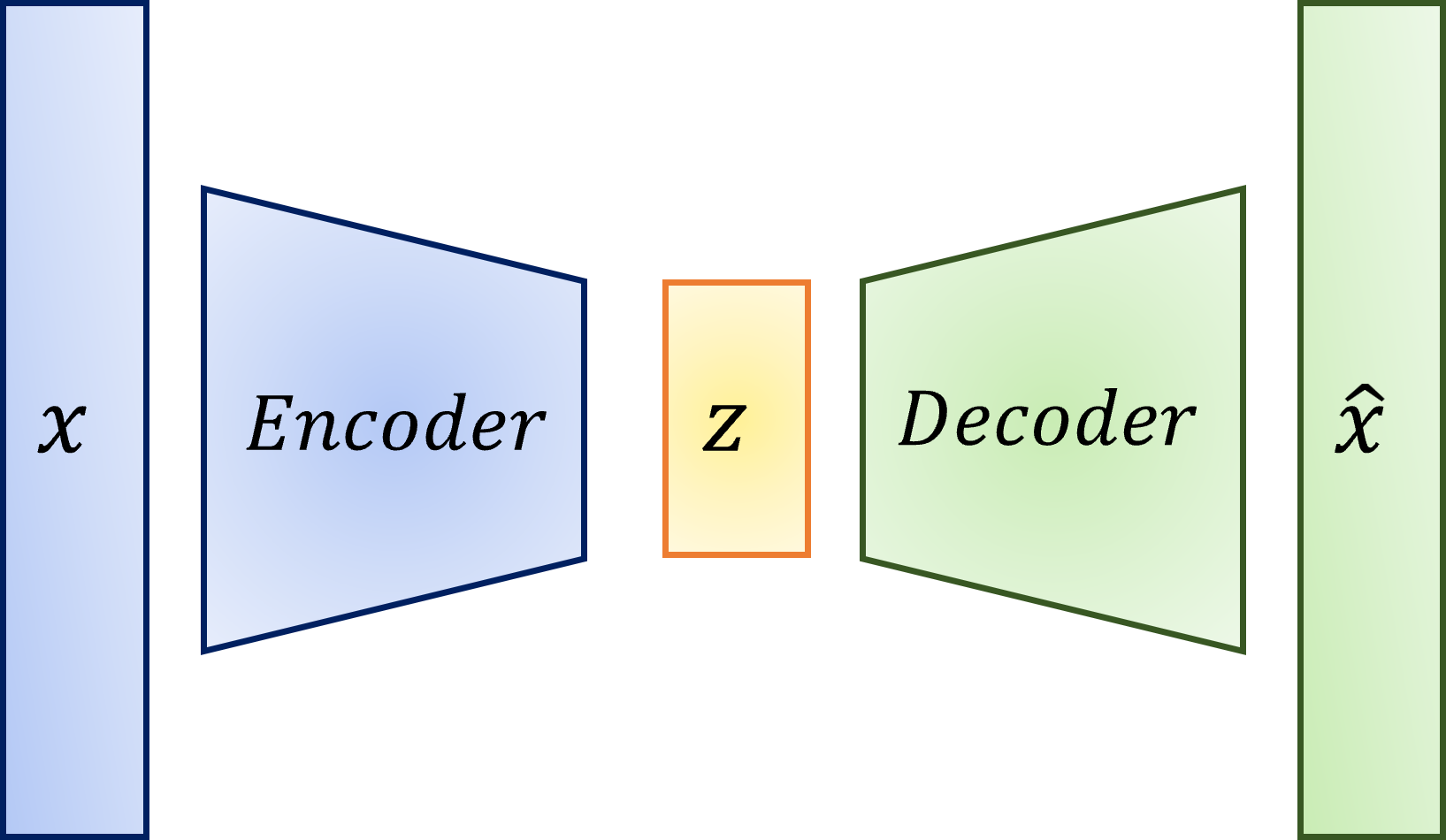}
    \caption{Variational autoencoder architecture}
    \label{fig:Figure_2}
\end{figure}

An autoencoder architecture is shown in Figure \ref{fig:Figure_2}. Here $x$ and $\hat{x}$ are the input and the reconstructed output, respectively. Given a dataset $\mathbf{X} = \{x^{(i)}\}_{i=1}^{N}$, where $N$ is the number of data points, the assumption is that it is generated by a random process $p_{\theta}$ involving the random variable $\mathbf{z}$, which is the latent space. The probabilistic encoder and probabilistic decoder are given by the distribution $p_{\theta}(z|x)$ and $p_{\theta}(x|z)$, respectively. However, $p_{\theta}(z|x)$ is intractable, hence we assume a surrogate $q_{\phi}(z|x)$ which is a Gaussian distribution with mean and standard deviation as $\mu$ and $\sigma$ respectively. 

\par The goal is to learn the conditional distribution \( p_{\theta}(z|x) \), which we obtain by minimizing the distance between the surrogate and the original distribution 
using the KL divergence, given by $D_{\text{KL}} \left( q_{\phi}(z|x^{(i)}) \,\|\, p_{\theta}(z|x^{(i)}) \right)$. \newline

Expanding the KL divergence term, we get

\begin{equation}
\log p_{\theta}(x^{(i)}) = D_{KL} \left( q_{\phi}(z|x^{(i)}) \parallel p_{\theta}(z|x^{(i)}) \right) + \mathcal{L}(\theta, \phi; x^{(i)})
\end{equation} 

The second part of the right-hand side of the equation is the evidence lower bound (ELBO) given by:

\begin{equation}
\mathcal{L}(\theta, \phi; x^{(i)}) = - D_{KL}(q_{\phi}(z|x^{(i)}) \parallel p_{\theta}(z)) + \mathbb{E}_{q_\phi (z|x^{(i)})} \left[ \log p_\theta (x^{(i)}|z) \right]
\end{equation}

$q_{\phi}(z|x^{(i)})$ is assumed to be a multivariate Gaussian with mean and standard deviation of $\mu$ and $\sigma$, and the prior $p_{\theta}(z)$ is assumed to be a multivariate Gaussian with zero mean and identity covariance. A detailed mathematical derivation can be found in the works by Kingma et al. \cite{Kingma2013} and Odaibo et al. \cite{odaibo2019tutorial}. The final loss function for a VAE is given by 

\begin{equation}
\mathcal{L}(\theta, \phi; x^{(i)}) \approx \frac{1}{2}\sum_{j=1}^{J} \left( 1 + \log \left( \left( \sigma_j^{(i)} \right)^2 \right) - \left( \mu_j^{(i)} \right)^2 - \left( \sigma_j^{(i)} \right)^2 \right) + \frac{1}{L}\sum_{l=1}^{L} \log p_{\theta}(x^{(i)}|z^{(i,l)})
\label{eq:elbo}
\end{equation}

Where $z^{(i,l)} = \mu^{(i)} + \sigma^{(i)} \odot \epsilon^{l}$ and $\epsilon^{l} \sim \mathcal{N}(0,I)$ is added as a part of the reparameterization trick to address the challenge of backpropagating through random variables \cite{Kingma2013}. In practice, $\mu$ and $\sigma$ are found using a multi-layered perceptron (MLP) network, where appropriate weights and biases are learned during training.

\section{Results - Effect of Input Configuration and Sequence Length} \label{appendixC}
\setcounter{table}{0}      
\setcounter{figure}{0}     

\begin{table}[H]
\centering
\scriptsize
\setlength{\tabcolsep}{3pt}
\begin{tabular}{@{}cc|ccccccc@{}}
\toprule
\textbf{Length} & \textbf{Input} & \multicolumn{7}{c}{\textbf{Time point}} \\
\cmidrule(l){3-9}
 & & \textbf{1} & \textbf{2} & \textbf{3} & \textbf{4} & \textbf{5} & \textbf{6} & \textbf{7} \\
\midrule
1 & 0 & \textbf{0.85$\pm$0.04} & 1.12$\pm$0.11 & 1.33$\pm$0.21 & 1.35$\pm$0.26 & 1.75$\pm$0.30 & 1.83$\pm$0.36 & 2.13$\pm$0.23 \\
2 & 0,1 & --- & \textbf{0.88$\pm$0.06} & 1.05$\pm$0.15 & 1.15$\pm$0.16 & 1.53$\pm$0.19 & 1.52$\pm$0.27 & 1.74$\pm$0.18 \\
3 & 0,1,2 & --- & --- & \textbf{0.87$\pm$0.08} & 0.97$\pm$0.14 & 1.41$\pm$0.26 & 1.38$\pm$0.25 & 1.55$\pm$0.14 \\
4 & 0,1,2,3 & --- & --- & --- & \textbf{0.85$\pm$0.16} & 1.16$\pm$0.27 & 1.25$\pm$0.23 & 1.33$\pm$0.24 \\
5 & 0,1,2,3,4 & --- & --- & --- & --- & \textbf{1.03$\pm$0.21} & 1.22$\pm$0.28 & 1.20$\pm$0.19 \\
6 & 0,1,2,3,4,5 & --- & --- & --- & --- & --- & \textbf{1.07$\pm$0.14} & \textbf{1.20$\pm$0.20} \\
\bottomrule
\end{tabular}
\caption{CDRSB prediction results with varying input lengths}
\label{tab:CDRSB_input_length}
\end{table}

\begin{table}[H]
\centering
\scriptsize
\setlength{\tabcolsep}{3pt}
\begin{tabular}{@{}c|cccccccc@{}}
\toprule
\textbf{Inputs$\downarrow$/Target$\rightarrow$} & \textbf{0} & \textbf{1} & \textbf{2} & \textbf{3} & \textbf{4} & \textbf{5} & \textbf{6} & \textbf{7} \\
\midrule
0 & --- & 0.85 & 1.13 & 1.34 & 1.37 & 1.74 & 1.79 & 2.11 \\
1 & \textbf{0.51} & --- & \textbf{0.85} & 1.05 & 1.16 & 1.58 & 1.53 & 1.76 \\
2 & 0.60 & \textbf{0.83} & --- & \textbf{0.88} & 0.97 & 1.40 & 1.39 & 1.62 \\
3 & 0.63 & 0.95 & 1.05 & --- & \textbf{0.92} & 1.22 & 1.25 & 1.39 \\
4 & 0.63 & 0.99 & 1.19 & 1.21 & --- & \textbf{1.13} & 1.24 & 1.35 \\
5 & 0.65 & 1.04 & 1.29 & 1.42 & 1.24 & --- & \textbf{1.10} & 1.38 \\
6 & 0.66 & 1.05 & 1.33 & 1.46 & 1.38 & 1.42 & --- & \textbf{1.08} \\
7 & 0.64 & 1.06 & 1.31 & 1.43 & 1.40 & 1.51 & 1.32 & --- \\
\bottomrule
\end{tabular}
\caption{CDRSB prediction results for different input/target time point combinations}
\label{tab:CDRSB_input_target}
\end{table}

\begin{table}[H]
\centering
\scriptsize
\setlength{\tabcolsep}{3pt}
\begin{tabular}{@{}cc|ccccccc@{}}
\toprule
\textbf{Length} & \textbf{Input} & \multicolumn{7}{c}{\textbf{Time point}} \\
\cmidrule(l){3-9}
 & & \textbf{1} & \textbf{2} & \textbf{3} & \textbf{4} & \textbf{5} & \textbf{6} & \textbf{7} \\
\midrule
1 & 0 & \textbf{1.86$\pm$0.06} & 2.12$\pm$0.19 & 2.19$\pm$0.14 & 2.30$\pm$0.22 & 2.70$\pm$0.27 & 2.63$\pm$0.43 & 2.72$\pm$0.21 \\
2 & 0,1 & --- & \textbf{1.76$\pm$0.09} & 1.88$\pm$0.08 & 2.00$\pm$0.17 & 2.40$\pm$0.27 & 2.42$\pm$0.58 & 2.40$\pm$0.29 \\
3 & 0,1,2 & --- & --- & \textbf{1.72$\pm$0.17} & 1.85$\pm$0.03 & 2.23$\pm$0.19 & 2.23$\pm$0.31 & 2.27$\pm$0.53 \\
4 & 0,1,2,3 & --- & --- & --- & \textbf{1.67$\pm$0.05} & 2.17$\pm$0.13 & 2.04$\pm$0.31 & 2.12$\pm$0.34 \\
5 & 0,1,2,3,4 & --- & --- & --- & --- & \textbf{2.11$\pm$0.31} & \textbf{2.04$\pm$0.32} & \textbf{2.12$\pm$0.30} \\
6 & 0,1,2,3,4,5 & --- & --- & --- & --- & --- & 2.24$\pm$0.38 & 2.28$\pm$0.45 \\
\bottomrule
\end{tabular}
\caption{MMSE prediction results with varying input lengths}
\label{tab:MMSE_input_length}
\end{table}

\begin{table}[H]
\centering
\scriptsize
\setlength{\tabcolsep}{3pt}
\begin{tabular}{@{}c|cccccccc@{}}
\toprule
\textbf{Inputs$\downarrow$/Target$\rightarrow$} & \textbf{0} & \textbf{1} & \textbf{2} & \textbf{3} & \textbf{4} & \textbf{5} & \textbf{6} & \textbf{7} \\
\midrule
0 & --- & 1.89 & 2.08 & 2.26 & 2.41 & 2.81 & 2.63 & 2.86 \\
1 & \textbf{1.35} & --- & \textbf{1.84} & 2.07 & 2.24 & 2.70 & 2.66 & 2.84 \\
2 & 1.48 & \textbf{1.81} & --- & \textbf{1.84} & 2.10 & 2.44 & 2.42 & 2.45 \\
3 & 1.46 & 1.90 & 2.05 & --- & \textbf{1.86} & 2.38 & 2.30 & 2.54 \\
4 & 1.46 & 1.94 & 2.15 & 2.01 & --- & \textbf{2.04} & \textbf{2.08} & 2.27 \\
5 & 1.53 & 1.99 & 2.29 & 2.38 & 2.25 & --- & 2.22 & 2.35 \\
6 & 1.49 & 1.98 & 2.24 & 2.33 & 2.33 & 2.47 & --- & \textbf{2.06} \\
7 & 1.51 & 2.01 & 2.26 & 2.39 & 2.40 & 2.65 & 2.45 & --- \\
\bottomrule
\end{tabular}
\caption{MMSE prediction results for different input/target time point combinations}
\label{tab:MMSE_input_target}
\end{table}

\begin{table}[H]
\centering
\scriptsize
\setlength{\tabcolsep}{3pt}
\begin{tabular}{@{}c|cccccccc@{}}
\toprule
\textbf{Inputs$\downarrow$/Target$\rightarrow$} & \textbf{0} & \textbf{1} & \textbf{2} & \textbf{3} & \textbf{4} & \textbf{5} & \textbf{6} & \textbf{7} \\
\midrule
0, 1 & --- & --- & 0.85 & 1.06 & 1.15 & 1.55 & 1.50 & 1.79 \\
0, 2 & --- & 0.73 & --- & 0.89 & 1.01 & 1.40 & 1.36 & 1.60 \\
0, 3 & --- & 0.78 & 0.91 & --- & 0.86 & 1.18 & 1.25 & 1.47 \\
0, 4 & --- & 0.83 & 1.04 & 1.09 & --- & 1.15 & 1.32 & 1.45 \\
0, 5 & --- & 0.83 & 1.09 & 1.24 & 1.14 & --- & 1.13 & 1.42 \\
0, 6 & --- & 0.84 & 1.10 & 1.27 & 1.23 & 1.33 & --- & 1.01 \\
0, 7 & --- & 0.83 & 1.09 & 1.25 & 1.28 & 1.44 & 1.30 & --- \\
1, 2 & 0.50 & --- & --- & 0.84 & 0.97 & 1.43 & 1.47 & 1.64 \\
1, 3 & 0.52 & --- & 0.77 & --- & 0.81 & 1.11 & 1.17 & 1.23 \\
1, 4 & 0.51 & --- & 0.80 & 0.84 & --- & 1.03 & 1.18 & 1.29 \\
1, 5 & 0.51 & --- & 0.85 & 0.97 & 0.96 & --- & 1.01 & 1.20 \\
1, 6 & 0.51 & --- & 0.84 & 0.99 & 1.08 & 1.28 & --- & 1.02 \\
1, 7 & 0.51 & --- & 0.85 & 0.99 & 1.07 & 1.33 & 1.16 & --- \\
2, 3 & 0.58 & 0.82 & --- & --- & 0.82 & 1.15 & 1.24 & 1.32 \\
2, 4 & 0.58 & 0.81 & --- & 0.79 & --- & 0.96 & 1.14 & 1.19 \\
2, 5 & 0.58 & 0.82 & --- & 0.83 & 0.79 & --- & 0.96 & 1.14 \\
2, 6 & 0.58 & 0.81 & --- & 0.83 & 0.90 & 1.09 & --- & 1.06 \\
2, 7 & 0.58 & 0.82 & --- & 0.81 & 0.86 & 1.14 & 1.10 & --- \\
3, 4 & 0.61 & 0.95 & 1.04 & --- & --- & 1.00 & 1.14 & 1.20 \\
3, 5 & 0.62 & 0.94 & 1.04 & --- & 0.78 & --- & 0.94 & 1.09 \\
3, 6 & 0.62 & 0.94 & 1.04 & --- & 0.89 & 1.03 & --- & 0.96 \\
3, 7 & 0.62 & 0.96 & 1.04 & --- & 0.88 & 1.09 & 1.01 & --- \\
4, 5 & 0.64 & 0.99 & 1.19 & 1.15 & --- & --- & 1.02 & 1.16 \\
4, 6 & 0.64 & 1.00 & 1.19 & 1.16 & --- & 0.90 & --- & 0.88 \\
4, 7 & 0.65 & 1.00 & 1.19 & 1.16 & --- & 1.00 & 1.01 & --- \\
5, 6 & 0.66 & 1.05 & 1.30 & 1.41 & 1.22 & --- & --- & 1.01 \\
5, 7 & 0.66 & 1.04 & 1.27 & 1.35 & 1.16 & --- & 0.87 & --- \\
6, 7 & 0.65 & 1.06 & 1.32 & 1.43 & 1.35 & 1.42 & --- & --- \\
\bottomrule
\end{tabular}
\caption{CDRSB prediction results for different two-input/target time point combinations}
\label{tab:CDRSB_two_inputs}
\end{table}

\begin{table}[H]
\centering
\scriptsize
\setlength{\tabcolsep}{3pt}
\begin{tabular}{@{}c|cccccccc@{}}
\toprule
\textbf{Inputs$\downarrow$/Target$\rightarrow$} & \textbf{0} & \textbf{1} & \textbf{2} & \textbf{3} & \textbf{4} & \textbf{5} & \textbf{6} & \textbf{7} \\
\midrule
0, 1 & --- & --- & 1.76 & 1.88 & 2.00 & 2.40 & 2.34 & 2.40 \\
0, 2 & --- & 1.67 & --- & 1.81 & 2.02 & 2.39 & 2.50 & 2.45 \\
0, 3 & --- & 1.84 & 1.95 & --- & 1.90 & 2.38 & 2.29 & 2.41 \\
0, 4 & --- & 1.82 & 1.98 & 1.87 & --- & 2.16 & 2.01 & 2.26 \\
0, 5 & --- & 1.83 & 2.03 & 2.08 & 2.04 & --- & 2.06 & 2.32 \\
0, 6 & --- & 1.85 & 2.06 & 2.07 & 2.08 & 2.24 & --- & 1.97 \\
0, 7 & --- & 1.86 & 2.08 & 2.14 & 2.27 & 2.53 & 2.27 & --- \\
1, 2 & 1.31 & --- & --- & 1.64 & 1.79 & 2.14 & 2.21 & 2.23 \\
1, 3 & 1.37 & --- & 1.63 & --- & 1.68 & 2.09 & 2.24 & 2.38 \\
1, 4 & 1.29 & --- & 1.66 & 1.65 & --- & 1.82 & 2.07 & 2.21 \\
1, 5 & 1.30 & --- & 1.73 & 1.81 & 1.75 & --- & 1.96 & 2.18 \\
1, 6 & 1.33 & --- & 1.76 & 1.89 & 1.95 & 2.15 & --- & 2.00 \\
1, 7 & 1.44 & --- & 1.93 & 2.11 & 2.29 & 2.75 & 2.49 & --- \\
2, 3 & 1.39 & 1.76 & --- & --- & 1.70 & 2.11 & 2.08 & 2.23 \\
2, 4 & 1.38 & 1.72 & --- & 1.53 & --- & 1.76 & 1.96 & 2.07 \\
2, 5 & 1.40 & 1.75 & --- & 1.66 & 1.79 & --- & 2.02 & 2.18 \\
2, 6 & 1.38 & 1.73 & --- & 1.65 & 1.79 & 1.96 & --- & 1.84 \\
2, 7 & 1.47 & 1.81 & --- & 1.83 & 2.00 & 2.29 & 2.19 & --- \\
3, 4 & 1.49 & 1.91 & 2.03 & --- & --- & 1.94 & 1.94 & 2.12 \\
3, 5 & 1.42 & 1.87 & 2.02 & --- & 1.73 & --- & 1.99 & 2.21 \\
3, 6 & 1.46 & 1.90 & 2.06 & --- & 1.77 & 2.13 & --- & 2.14 \\
3, 7 & 1.46 & 1.89 & 2.04 & --- & 1.80 & 2.24 & 2.02 & --- \\
4, 5 & 1.50 & 1.97 & 2.22 & 2.19 & --- & --- & 2.12 & 2.36 \\
4, 6 & 1.47 & 1.94 & 2.13 & 2.01 & --- & 1.85 & --- & 1.79 \\
4, 7 & 1.46 & 1.90 & 2.09 & 1.94 & --- & 1.90 & 1.72 & --- \\
5, 6 & 1.49 & 1.98 & 2.20 & 2.31 & 2.15 & --- & --- & 2.03 \\
5, 7 & 1.49 & 1.97 & 2.23 & 2.28 & 2.11 & --- & 1.97 & --- \\
6, 7 & 1.49 & 1.97 & 2.22 & 2.31 & 2.23 & 2.27 & --- & --- \\
\bottomrule
\end{tabular}
\caption{MMSE prediction results for different two-input/target time point combinations}
\label{tab:MMSE_two_inputs}
\end{table}

\section{Ablation Results Tables} \label{appendixD}
\setcounter{table}{0}
\setcounter{figure}{0}

The tables below present the ablation results for predictions of CDR-SB and MMSE scores. 

\begin{table}[ht]
  \centering
  \scriptsize
  \renewcommand{\arraystretch}{0.9}  
  \setlength{\tabcolsep}{2pt}  
  \begin{adjustbox}{width=\textwidth}
    \begin{tabular}{l*{10}{c}}  
      \toprule
      & \multicolumn{9}{c}{\textbf{Time points}} & \\
      \cmidrule(lr){2-10}
      \textbf{Model} & \textbf{2}   & \textbf{3}   & \textbf{4}   & \textbf{5}   & \textbf{6}   & \textbf{7}   & \textbf{8}   & \textbf{9}   & \textbf{10}  & \textbf{Avg}  \\
      \midrule
      Our Model & 0.8775 & \textbf{1.0519} & 1.1476 & \textbf{1.5342} & 1.5182 & \textbf{1.7380} & 1.7151 & 2.2122 & 2.1530 & \textbf{1.4464} \\
      Avg hidden state & 0.8519 & 1.0596 & 1.1783 & 1.5632 & 1.4931 & 1.7813 & 1.6866 & 2.2351 & 2.1921 & 1.4491 \\
      Attention scores & 0.8543 & 1.0718 & 1.1744 & 1.5761 & 1.4906 & 1.7701 & 1.6968 & 2.2642 & 2.2259 & 1.4591 \\
      \midrule
      W/O Age & 0.8725 & 1.0874 & 1.1982 & 1.6321 & 1.5827 & 1.9032 & 1.7214 & 2.2672 & 2.2550 & 1.4891 \\
      W/O Education & \textbf{0.8460} & 1.0590 & 1.1527 & 1.5546 & 1.5017 & 1.7869 & 1.7059 & 2.2082 & 2.2261 & 1.4483 \\
      W/O BMI & 0.8699 & 1.0726 & 1.1716 & 1.6120 & 1.5323 & 1.8166 & 1.6984 & 2.1979 & 2.2126 & 1.4605 \\
      W/O Hypertension    & 0.8486 & 1.0608 & 1.1538 & 1.5419 & \textbf{1.4770} & 1.7808 & 1.6526 & 2.1442 & 2.0515 & \textbf{1.4148} \\
      W/O Gender & 0.8575 & 1.0647 & \textbf{1.1414} & 1.5448 & 1.4966 & 1.7705 & \textbf{1.6828} & \textbf{2.1410} & \textbf{1.9949} & 1.4160 \\
      W/O MMSE & 0.8549 & 1.0671 & 1.1853 & 1.6119 & 1.5448 & 1.8638 & 1.7231 & 2.2463 & 2.1849 & 1.4673 \\
      W/O Diagnosis & 0.8717 & 1.0778 & 1.1912 & 1.6208 & 1.5292 & 1.7973 & 1.7031 & 2.1819 & 2.0867 & 1.4506 \\
      W/O APOE4 & 0.8575 & 1.0711 & 1.1979 & 1.5905 & 1.5692 & 1.7813 & 1.7471 & 2.3422 & 2.1448 & 1.4772 \\
      \bottomrule
    \end{tabular}
  \end{adjustbox}
  \caption{Mean Absolute Error comparison for different CDR–SB model variants and ablations}
  \label{tab:CDRSB_ablation}
\end{table}

\begin{table}[ht]
  \centering
  \scriptsize
  \renewcommand{\arraystretch}{0.9}
  \setlength{\tabcolsep}{2pt}
  \begin{adjustbox}{width=\textwidth}
    \begin{tabular}{l*{10}{c}}  
      \toprule
      & \multicolumn{9}{c}{\textbf{Time points}} & \\
      \cmidrule(lr){2-10}
      \textbf{Model} & \textbf{2}   & \textbf{3}   & \textbf{4}   & \textbf{5}   & \textbf{6}   & \textbf{7}   & \textbf{8}   & \textbf{9}   & \textbf{10}  & \textbf{Avg}  \\
      \midrule
      Our Model & 1.7553 & \textbf{1.8831} & 1.9974 & 2.3953 & \textbf{2.3411} & 2.3993 & 2.4463 & 3.1661 & 2.8090 & \textbf{2.2653} \\
      Avg hidden state & 1.8035 & 1.9238 & 2.0545 & 2.5065 & 2.4540 & 2.7445 & 2.4356 & 3.3219 & 3.4993 & 2.3957 \\
      Attention scores & 1.8727 & 2.0092 & 2.1837 & 2.5623 & 2.5861 & 2.5402 & 2.5364 & 3.2326 & 3.3331 & 2.4309 \\
      \midrule
      W/O Age & 1.9668 & 2.0471 & 2.2231 & 2.6650 & 2.5498 & 2.7079 & 2.4619 & 3.0591 & 2.8691 & 2.3982 \\
      W/O Education & 1.7758 & 1.8920 & 1.9951 & 2.4271 & 2.4263 & 2.5035 & \textbf{2.3318} & 2.9942 & 2.8277 & \textbf{2.2460} \\
      W/O BMI & 1.9006 & 1.9517 & 2.1259 & 2.5786 & 2.5384 & 2.7408 & 2.4247 & \textbf{2.9510} & 2.6786 & 2.3186 \\
      W/O Hypertension & 1.7923 & 1.9229 & 2.0394 & 2.5020 & 2.3750 & 2.4815 & 2.4687 & 3.1082 & 2.8432 & 2.2962 \\
      W/O Gender & 1.7958 & 1.9107 & 1.9995 & 2.3623 & 2.3633 & 2.4375 & 2.4559 & 3.1832 & 3.0838 & 2.3083 \\
      W/O CDR-SB  & 1.8773 & 1.9734 & 2.1403 & 2.5476 & 2.5193 & 2.7618 & 2.4198 & 2.9528 & \textbf{2.7399} & 2.3212 \\
      W/O Diagnosis & \textbf{1.7220} & 1.9053 & \textbf{1.9494} & \textbf{2.2861} & 2.3657 & \textbf{2.3734} & 2.5523 & 3.3700 & 3.3669 & 2.3411 \\
      W/O APOE4 & 1.8181 & 1.9104 & 2.0613 & 2.5082 & 2.6148 & 2.8063 & 2.6160 & 3.2708 & 3.0214 & 2.3880 \\
      \bottomrule
    \end{tabular}
  \end{adjustbox}
  \caption{Mean Absolute Error (MAE) comparison for different MMSE model variants and ablations}
  \label{tab:MMSE_ablation}
\end{table}

\bibliography{mybibfile}

@misc{Kingma2013,
   title={Auto-Encoding Variational Bayes}, 
   author={Diederik P Kingma and Max Welling},
   year={2022},
   eprint={1312.6114},
   archivePrefix={arXiv},
   primaryClass={stat.ML},
   url={https://arxiv.org/abs/1312.6114}
}

@article{Mueller2005,
   author = {Susanne G. Mueller and Michael W. Weiner and Leon J. Thal and Ronald C. Petersen and Clifford Jack and William Jagust and John Q. Trojanowski and Arthur W. Toga and Laurel Beckett},
   doi = {10.1016/j.nic.2005.09.008},
   issn = {10525149},
   issue = {4},
   journal = {Neuroimaging Clinics of North America},
   month = {11},
   pages = {869-877},
   title = {The Alzheimer's Disease Neuroimaging Initiative},
   volume = {15},
   year = {2005}
}

@article{Jeong2024,
   author = {Seungwoo Jeong and Wonsik Jung and Junghyo Sohn and Heung Il Suk},
   doi = {10.1109/TNNLS.2024.3394598},
   issn = {21622388},
   journal = {IEEE Transactions on Neural Networks and Learning Systems},
   publisher = {Institute of Electrical and Electronics Engineers Inc.},
   title = {Deep Geometric Learning With Monotonicity Constraints for Alzheimer's Disease Progression},
   year = {2024}
}

@misc{Chen2018,
      title={Neural Ordinary Differential Equations}, 
      author={Ricky T. Q. Chen and Yulia Rubanova and Jesse Bettencourt and David Duvenaud},
      year={2019},
      eprint={1806.07366},
      archivePrefix={arXiv},
      primaryClass={cs.LG},
      url={https://arxiv.org/abs/1806.07366}, 
}

@article{Liang2021,
   author = {Wei Liang and Kai Zhang and Peng Cao and Xiaoli Liu and Jinzhu Yang and Osmar Zaiane},
   doi = {10.1016/j.compbiomed.2021.104935},
   issn = {18790534},
   journal = {Computers in Biology and Medicine},
   month = {11},
   pmid = {34656869},
   publisher = {Elsevier Ltd},
   title = {Rethinking modeling Alzheimer's disease progression from a multi-task learning perspective with deep recurrent neural network},
   volume = {138},
   year = {2021}
}

@article{Liu2020,
   author = {Mingxia Liu and Jun Zhang and Chunfeng Lian and DInggang Shen},
   doi = {10.1109/TCYB.2019.2904186},
   issn = {21682275},
   issue = {7},
   journal = {IEEE Transactions on Cybernetics},
   month = {7},
   pages = {3381-3392},
   pmid = {30932861},
   publisher = {Institute of Electrical and Electronics Engineers Inc.},
   title = {Weakly Supervised Deep Learning for Brain Disease Prognosis Using MRI and Incomplete Clinical Scores},
   volume = {50},
   year = {2020}
}

@article{Morar2023,
   author = {Ulyana Morar and Harold Martin and P. M. Robin and Walter Izquierdo and Elaheh Zarafshan and Parisa Forouzannezhad and Elona Unger and Mercedes Cabrerizo and Rosie E. Curiel Cid and Monica Rosselli and Armando Barreto and Naphtali Rishe and David E. Vaillancourt and Steven T. DeKosky and David Loewenstein and Ranjan Duara and Malek Adjouadi},
   doi = {10.1007/s12559-023-10169-w},
   issn = {18669964},
   issue = {6},
   journal = {Cognitive Computation},
   month = {11},
   pages = {2062-2086},
   publisher = {Springer},
   title = {Prediction of Cognitive Test Scores from Variable Length Multimodal Data in Alzheimer’s Disease},
   volume = {15},
   year = {2023}
}

@article{Lei2022,
   author = {Baiying Lei and Enmin Liang and Mengya Yang and Peng Yang and Feng Zhou and Ee Leng Tan and Yi Lei and Chuan Ming Liu and Tianfu Wang and Xiaohua Xiao and Shuqiang Wang},
   doi = {10.1016/j.eswa.2021.115966},
   issn = {09574174},
   journal = {Expert Systems with Applications},
   month = {1},
   publisher = {Elsevier Ltd},
   title = {Predicting clinical scores for Alzheimer's disease based on joint and deep learning},
   volume = {187},
   year = {2022}
}

@article{Mukherji2022,
   author = {Devarshi Mukherji and Manibrata Mukherji and Nivedita Mukherji},
   doi = {10.1186/s40708-022-00169-1},
   issn = {21984026},
   issue = {1},
   journal = {Brain Informatics},
   month = {12},
   publisher = {Springer Science and Business Media Deutschland GmbH},
   title = {Early detection of Alzheimer’s disease using neuropsychological tests: a predict–diagnose approach using neural networks},
   volume = {9},
   year = {2022}
}

@article{Nguyen2020,
   author = {Minh Nguyen and Tong He and Lijun An and Daniel C. Alexander and Jiashi Feng and B. T.Thomas Yeo},
   doi = {10.1016/j.neuroimage.2020.117203},
   issn = {10959572},
   journal = {NeuroImage},
   month = {11},
   pmid = {32763427},
   publisher = {Academic Press Inc.},
   title = {Predicting Alzheimer's disease progression using deep recurrent neural networks},
   volume = {222},
   year = {2020}
}

@article{Devanarayan2024,
   author = {Viswanath Devanarayan and Yuanqing Ye and Arnaud Charil and Erica Andreozzi and Pallavi Sachdev and Daniel A. Llano and Lu Tian and Liang Zhu and Harald Hampel and Lynn Kramer and Shobha Dhadda and Michael Irizarry},
   doi = {10.1002/alz.13565},
   issn = {15525279},
   issue = {3},
   journal = {Alzheimer's and Dementia},
   month = {3},
   pages = {1725-1738},
   pmid = {38087949},
   publisher = {John Wiley and Sons Inc},
   title = {Predicting clinical progression trajectories of early Alzheimer's disease patients},
   volume = {20},
   year = {2024}
}

@article{Poonam2024,
   author = {Km Poonam and Rajlakshmi Guha and Partha P. Chakrabarti},
   doi = {10.1109/JBHI.2024.3386801},
   issn = {21682208},
   issue = {7},
   journal = {IEEE Journal of Biomedical and Health Informatics},
   pages = {4184-4193},
   pmid = {38593020},
   publisher = {Institute of Electrical and Electronics Engineers Inc.},
   title = {Predicting Alzheimer's Disease Progression Using a Versatile Sequence-Length-Adaptive Encoder-Decoder LSTM Architecture},
   volume = {28},
   year = {2024}
}

@article{Yuan2024,
   author = {Zengbei Yuan and Xinlin Li and Zezhou Hao and Zhixian Tang and Xufeng Yao and Tao Wu},
   doi = {10.1038/s41598-024-62712-w},
   issn = {20452322},
   issue = {1},
   journal = {Scientific Reports},
   month = {12},
   pmid = {38796518},
   publisher = {Nature Research},
   title = {Intelligent prediction of Alzheimer’s disease via improved multifeature squeeze-and-excitation-dilated residual network},
   volume = {14},
   year = {2024}
}

@article{Jung2021,
   author = {Wonsik Jung and Eunji Jun and Heung Il Suk and Alzheimer's Disease Neuroimaging Initiative},
   doi = {10.1016/j.neuroimage.2021.118143},
   issn = {10959572},
   journal = {NeuroImage},
   month = {8},
   pmid = {33991694},
   publisher = {Academic Press Inc.},
   title = {Deep recurrent model for individualized prediction of Alzheimer's disease progression},
   volume = {237},
   year = {2021}
}

@article{Lei2020,
   author = {Baiying Lei and Mengya Yang and Peng Yang and Feng Zhou and Wen Hou and Wenbin Zou and Xia Li and Tianfu Wang and Xiaohua Xiao and Shuqiang Wang},
   doi = {10.1016/j.patcog.2020.107247},
   issn = {00313203},
   journal = {Pattern Recognition},
   month = {6},
   publisher = {Elsevier Ltd},
   title = {Deep and joint learning of longitudinal data for Alzheimer's disease prediction},
   volume = {102},
   year = {2020}
}

@misc{Peterson2017,
      title={Personalized Gaussian Processes for Future Prediction of Alzheimer's Disease Progression}, 
      author={Kelly Peterson and Ognjen Rudovic and Ricardo Guerrero and Rosalind W. Picard},
      year={2018},
      eprint={1712.00181},
      archivePrefix={arXiv},
      primaryClass={cs.LG},
      url={https://arxiv.org/abs/1712.00181}, 
}

@article{Puri2022,
   author = {Chetanya Puri and Gerben Kooijman and Bart Vanrumste and Stijn Luca},
   doi = {10.1109/JBHI.2022.3214343},
   issn = {21682208},
   issue = {12},
   journal = {IEEE Journal of Biomedical and Health Informatics},
   month = {12},
   pages = {6126-6137},
   pmid = {36227825},
   publisher = {Institute of Electrical and Electronics Engineers Inc.},
   title = {Forecasting Time Series in Healthcare With Gaussian Processes and Dynamic Time Warping Based Subset Selection},
   volume = {26},
   year = {2022}
}

@article{Jiang2021,
   author = {Shu Jiang and Yijun Xie and Graham A. Colditz},
   doi = {10.1111/rssc.12449},
   issn = {14679876},
   issue = {1},
   journal = {Journal of the Royal Statistical Society. Series C: Applied Statistics},
   month = {1},
   pages = {66-79},
   publisher = {Blackwell Publishing Ltd},
   title = {Functional ensemble survival tree: Dynamic prediction of Alzheimer’s disease progression accommodating multiple time-varying covariates},
   volume = {70},
   year = {2021}
}

@inproceedings{Morar2020,
   author = {Ulyana Morar and Harold Martin and Walter Izquierdo and Parisa Forouzannezhad and Elaheh Zarafshan and Rosie E. Curiel and Monica Roselli and David Loewenstein and Ranjan Duara and Elona Unger and Malek Adjouadi},
   doi = {10.1109/CSCI51800.2020.00144},
   isbn = {9781728176246},
   booktitle = {Proceedings - 2020 International Conference on Computational Science and Computational Intelligence, CSCI 2020},
   month = {12},
   pages = {761-766},
   publisher = {Institute of Electrical and Electronics Engineers Inc.},
   title = {A Deep-Learning Approach for the Prediction of Mini-Mental State Examination Scores in a Multimodal Longitudinal Study},
   year = {2020}
}

@article{Tabarestani2020,
   author = {Solale Tabarestani and Maryamossadat Aghili and Mohammad Eslami and Mercedes Cabrerizo and Armando Barreto and Naphtali Rishe and Rosie E. Curiel and David Loewenstein and Ranjan Duara and Malek Adjouadi},
   doi = {10.1016/j.neuroimage.2019.116317},
   issn = {10959572},
   journal = {NeuroImage},
   month = {2},
   pmid = {31678502},
   publisher = {Academic Press Inc.},
   title = {A distributed multitask multimodal approach for the prediction of Alzheimer's disease in a longitudinal study},
   volume = {206},
   year = {2020}
}

@inproceedings{Poonam2023,
  author={Poonam, Km and Guha, Rajlakshmi and Chakrabarti, Partha P},
  booktitle={2023 45th Annual International Conference of the IEEE Engineering in Medicine \& Biology Society (EMBC)}, 
  title={Accurate Prediction of Alzheimer’s Disease Progression Trajectory via a Novel Encoder-Decoder LSTM Architecture}, 
  year={2023},
  volume={},
  number={},
  pages={1-4},
  doi={10.1109/EMBC40787.2023.10340517}}

@article{Mouchet2021,
   author = {Julie Mouchet and Keith A. Betts and Mihaela V. Georgieva and Raluca Ionescu-Ittu and Lesley M. Butler and Xavier Teitsma and Paul Delmar and Thomas Kulalert and JingJing Zhu and Neema Lema and Urvi Desai},
   doi = {10.3233/JAD-210305},
   issn = {13872877},
   issue = {4},
   journal = {Journal of Alzheimer's Disease},
   month = {8},
   pages = {1667-1682},
   title = {Classification, Prediction, and Concordance of Cognitive and Functional Progression in Patients with Mild Cognitive Impairment in the United States: A Latent Class Analysis},
   volume = {82},
   year = {2021}
}

@article{Folstein1975,
   author = {Marshal F. Folstein and Susan E. Folstein and Paul R. McHugh},
   doi = {10.1016/0022-3956(75)90026-6},
   issn = {00223956},
   issue = {3},
   journal = {Journal of Psychiatric Research},
   month = {11},
   pages = {189-198},
   title = {“Mini-mental state”: A practical method for grading the cognitive state of patients for the clinician},
   volume = {12},
   year = {1975}
}

@article{Tzeng2022,
   author = {Ray-Chang Tzeng and Yu-Wan Yang and Kai-Cheng Hsu and Hsin-Te Chang and Pai-Yi Chiu},
   doi = {10.3389/fnagi.2022.1021792},
   issn = {1663-4365},
   journal = {Frontiers in Aging Neuroscience},
   month = {9},
   title = {Sum of boxes of the clinical dementia rating scale highly predicts conversion or reversion in predementia stages},
   volume = {14},
   year = {2022}
}

@article{Lowe2024,
   author = {Val J. Lowe and Carly T. Mester and Emily S. Lundt and Jeyeon Lee and Sujala Ghatamaneni and Alicia Algeciras‐Schimnich and Michelle R. Campbell and Jonathan Graff‐Radford and Aivi Nguyen and Hoon‐Ki Min and Matthew L. Senjem and Mary M. Machulda and Christopher G. Schwarz and Dennis W. Dickson and Melissa E. Murray and Karunya K. Kandimalla and Kejal Kantarci and Bradley Boeve and Prashanthi Vemuri and David T. Jones and David Knopman and Clifford R. Jack and Ronald C. Petersen and Michelle M. Mielke},
   doi = {10.1002/alz.14317},
   issn = {1552-5260},
   issue = {11},
   journal = {Alzheimer's \& Dementia},
   month = {11},
   pages = {8097-8112},
   title = {Amyloid PET detects the deposition of brain A$\beta$ earlier than CSF fluid biomarkers},
   volume = {20},
   year = {2024}
}

@article{Schaap2024,
   author = {Tamar Schaap and Pamela Thropp and Duygu Tosun},
   doi = {10.1002/alz.14306},
   issn = {15525279},
   journal = {Alzheimer's and Dementia},
   month = {12},
   pmid = {39428963},
   publisher = {John Wiley and Sons Inc},
   title = {Timing of Alzheimer's disease biomarker progressions: A two-decade observational study from the Alzheimer's Disease Neuroimaging Initiative (ADNI)},
   year = {2024}
}

@article{Blennow2017,
   author = {Kaj Blennow},
   doi = {10.1007/s40120-017-0073-9},
   issn = {2193-8253},
   issue = {S1},
   journal = {Neurology and Therapy},
   month = {7},
   pages = {15-24},
   title = {A Review of Fluid Biomarkers for Alzheimer’s Disease: Moving from CSF to Blood},
   volume = {6},
   year = {2017}
}

@misc{who_dementia_2024,
  author       = {{World Health Organization}},
  title        = {Dementia},
  year         = {2024},
  url          = {https://www.who.int/news-room/fact-sheets/detail/dementia},
  note         = {Accessed: 2025-05-19}
}

@article{Evans2019,
   author = {Stephanie Evans and Kevin McRae‐McKee and Christoforos Hadjichrysanthou and Mei Mei Wong and David Ames and Oscar Lopez and Frank de Wolf and Roy M. Anderson},
   doi = {10.1016/j.trci.2019.04.005},
   issn = {2352-8737},
   issue = {1},
   journal = {Alzheimer's \& Dementia: Translational Research \& Clinical Interventions},
   month = {1},
   pages = {515-523},
   title = {Alzheimer's disease progression and risk factors: A standardized comparison between six large data sets},
   volume = {5},
   year = {2019}
}

@misc{Cho2014,
      title={Learning Phrase Representations using RNN Encoder-Decoder for Statistical Machine Translation}, 
      author={Kyunghyun Cho and Bart van Merrienboer and Caglar Gulcehre and Dzmitry Bahdanau and Fethi Bougares and Holger Schwenk and Yoshua Bengio},
      year={2014},
      eprint={1406.1078},
      archivePrefix={arXiv},
      primaryClass={cs.CL},
      url={https://arxiv.org/abs/1406.1078}, 
}

@misc{Kendall2017,
   title={Auto-Encoding Variational Bayes}, 
   author={Diederik P Kingma and Max Welling},
   year={2022},
   eprint={1312.6114},
   archivePrefix={arXiv},
   primaryClass={stat.ML},
   url={https://arxiv.org/abs/1312.6114}
}

@book{Goodfellow2016,
    title={Deep Learning},
    author={Ian Goodfellow and Yoshua Bengio and Aaron Courville},
    publisher={MIT Press},
    url={http://www.deeplearningbook.org},
    year={2016}
}

@inproceedings{akiba2019optuna,
  title={Optuna: A next-generation hyperparameter optimization framework},
  author={Akiba, Takuya and Sano, Shotaro and Yanase, Toshihiko and Ohta, Takeru and Koyama, Masanori},
  booktitle={Proceedings of the 25th ACM SIGKDD international conference on knowledge discovery \& data mining},
  pages={2623--2631},
  year={2019}
}

@article{Zhang2024,
   author = {Jifa Zhang and Yinglu Zhang and Jiaxing Wang and Yilin Xia and Jiaxian Zhang and Lei Chen},
   doi = {10.1038/s41392-024-01911-3},
   issn = {2059-3635},
   issue = {1},
   journal = {Signal Transduction and Targeted Therapy},
   month = {8},
   pages = {211},
   title = {Recent advances in Alzheimer’s disease: mechanisms, clinical trials and new drug development strategies},
   volume = {9},
   year = {2024}
}

@article{odaibo2019tutorial,
  title={Tutorial: Deriving the standard variational autoencoder (vae) loss function},
  author={Odaibo, Stephen},
  journal={arXiv preprint arXiv:1907.08956},
  year={2019}
}

@article{Das,
title = {An interpretable Bayesian framework for Alzheimer’s disease prediction with uncertainty quantification},
journal = {Neuroscience},
volume = {589},
pages = {150-160},
year = {2025},
issn = {0306-4522},
doi = {https://doi.org/10.1016/j.neuroscience.2025.10.021},
url = {https://www.sciencedirect.com/science/article/pii/S0306452225010139},
author = {Ratnadeep Das and Atri Chatterjee and Sitikantha Roy}
}

@article{chung2014empirical,
  title={Empirical evaluation of gated recurrent neural networks on sequence modeling},
  author={Chung, Junyoung and Gulcehre, Caglar and Cho, KyungHyun and Bengio, Yoshua},
  journal={arXiv preprint arXiv:1412.3555},
  year={2014}
}

@article{FRANCOMARINA201072,
title = {The Mini-mental State Examination revisited: ceiling and floor effects after score adjustment for educational level in an aging Mexican population},
journal = {International Psychogeriatrics},
year = {2010},
issn = {1041-6102},
doi = {https://doi.org/10.1017/S1041610209990822},
author = {Francisco Franco-Marina and Jose Juan García-González and Fernando Wagner-Echeagaray and Joseph Gallo and Oscar Ugalde and Sergio Sánchez-García and Claudia Espinel-Bermúdez and Teresa Juárez-Cedillo and Miguel Ángel Villa Rodríguez and Carmen García-Peña},
}

@article{sando2008apoe,
  title={APOE $\varepsilon$4 lowers age at onset and is a high risk factor for Alzheimer's disease; A case control study from central Norway},
  author={Sando, Sigrid B and Melquist, Stacey and Cannon, Ashley and Hutton, Michael L and Sletvold, Olav and Saltvedt, Ingvild and White, Linda R and Lydersen, Stian and Aasly, Jan O},
  journal={BMC neurology},
  volume={8},
  number={1},
  pages={9},
  year={2008},
  publisher={Springer}
}

@article{roses2006discovery,
  title={On the discovery of the genetic association of Apolipoprotein E genotypes and common late-onset Alzheimer disease},
  author={Roses, Allen D},
  journal={Journal of Alzheimer’s Disease},
  volume={9},
  number={s3},
  pages={361--366},
  year={2006},
  publisher={SAGE Publications Sage UK: London, England}
}

@article{kim2009role,
  title={The role of apolipoprotein E in Alzheimer's disease},
  author={Kim, Jungsu and Basak, Jacob M and Holtzman, David M},
  journal={Neuron},
  volume={63},
  number={3},
  pages={287--303},
  year={2009},
  publisher={Elsevier}
}

@article{cho2022association,
  title={Association of late-life body mass index with the risk of Alzheimer disease: a 10-year nationwide population-based cohort study},
  author={Cho, Su Hwan and Jang, Minseol and Ju, Hyorim and Kang, Min Ju and Yun, Jae Moon and Yun, Jae Won},
  journal={Scientific Reports},
  volume={12},
  number={1},
  pages={15298},
  year={2022},
  publisher={Nature Publishing Group UK London}
}

@article{xie2016distinct,
  title={Distinct patterns of cognitive aging modified by education level and gender among adults with limited or no formal education: a normative study of the mini-mental state examination},
  author={Xie, Haiqun and Zhang, Chengguo and Wang, Yukai and Huang, Shuyun and Cui, Wei and Yang, Wenbin and Koski, Lisa and Xu, Xiping and Li, Youbao and Zheng, Meili and others},
  journal={Journal of alzheimer’s disease},
  volume={49},
  number={4},
  pages={961--969},
  year={2016},
  publisher={SAGE Publications Sage UK: London, England}
}

@article{chen2020learning,
  title={Learning neural event functions for ordinary differential equations},
  author={Chen, Ricky TQ and Amos, Brandon and Nickel, Maximilian},
  journal={arXiv preprint arXiv:2011.03902},
  year={2020}
}

\end{document}